%% file: paper.tex
\documentclass[letterpaper]{article} %
\usepackage{aaai2026}  %
\usepackage{times}  %
\usepackage{helvet}  %
\usepackage{courier}  %
\usepackage[hyphens]{url}  %
\usepackage{graphicx} %
\usepackage{natbib}  %
\usepackage{caption} %
\usepackage{amsfonts}
\usepackage{amssymb}
\usepackage{mathtools}
\usepackage{array}
\usepackage{booktabs}
\usepackage{xspace}
\usepackage{microtype}
\usepackage{verbatim}
\usepackage{capt-of}
\usepackage[main=american]{babel}
\usepackage[ruled,noline,noend]{algorithm2e}

\makeatletter
\g@addto@macro{\UrlBreaks}{\do\/\do\.\do\-\do\_\do\?\do\=\do\&}
\makeatother

\newcolumntype{H}{>{\setbox0=\hbox\bgroup}c<{\egroup}@{}}

\makeatletter
\renewcommand{\paragraph}{%
  \@startsection{paragraph}{4}{\z@}%
                {0.5ex \@plus 0.5ex \@minus 0.2ex}%
                {-1em}%
                {\normalsize\bf}%
}
\makeatother

\let\phi\varphi
\let\oldpar\paragraph
\renewcommand{\paragraph}[1]{\oldpar{#1.}}

\useshorthands*{"}
\defineshorthand{"-}{\babelhyphen{hard}}

\newcommand{\inlinecite}[1]{\citet{#1}}
\newcommand{\egcite}[1]{\citep[e.g.,][]{#1}}
\newcommand{\denselist}{\itemsep 0pt\partopsep 0pt}

\newcommand{\tuple}[1]{\ensuremath{\langle #1 \rangle}}
\newcommand{\eqdef}{\coloneqq}
\DeclareMathOperator{\dom}{dom}
\DeclareMathOperator{\pre}{pre}
\DeclareMathOperator{\eff}{eff}

\newcommand{\evBlMedTwo}{\texttt{evolved-blind-medium-2}\xspace}
\newcommand{\evFfNoneThree}{\texttt{evolved-ff-none-3}\xspace}
\newcommand{\evBlNoneThree}{\texttt{evolved-blind-none-3}\xspace}
\newcommand{\evBlMedConf}{\texttt{evolved-blind-medium-conf}\xspace}

\title{LLM-Evolved Domain-Independent Heuristics for Symbolic AI Planning}

\author{
    Elliot Gestrin,
    Jendrik Seipp
}
\affiliations{
    Link\"oping University, Sweden \\
    \{elliot.gestrin, jendrik.seipp\}@liu.se
}

\nocopyright

\begin{document}

\maketitle

\begin{abstract}
  Heuristic search is the dominant paradigm in symbolic AI planning, and the strongest heuristics are the result of decades of work by planning researchers.
  Recent work has shown that large language models (LLMs) can design heuristics for individual planning domains, but no LLM-generated heuristic has so far worked on arbitrary planning tasks.
  In this paper, we use evolutionary search to produce the first LLM-generated domain-independent heuristics that exceed the hand-engineered state of the art.
  We let an LLM mutate parent heuristics written in C++, store candidates in a MAP-Elites archive keyed on informedness and speed and calculate fitness scores by blending coverage with solving time.
  To place the evolved programs in context, we additionally benchmark a broad set of hand-engineered heuristics on their informedness--speed tradeoff, which to our knowledge has not been done before.
  On unseen testing domains, our best evolved heuristic solves more tasks than even the strongest baseline, with our full heuristic suite spanning the Pareto frontier of said tradeoff.
  We also find that seeding evolution from the trivial blind heuristic outperforms seeding from the strong FF heuristic, even when the resulting program is itself an FF variant, and that LLM reasoning effort affects how often candidates compile much more than the quality of those that do.
  Because the evolved programs are plain C++, they slot into existing planners as drop-in replacements and inherit the soundness and completeness guarantees of the underlying search.
\end{abstract}

\thanks{Accepted at the LM4Plan workshop at ICAPS 2026.}

\section{Introduction}

Symbolic AI planning underpins applications such as logistics, scheduling, robotics and formal verification, where an agent must compose long action sequences to reach a goal under hard correctness guarantees.
Given a symbolic world model, an initial state and a goal description, the task is to find a sequence of actions that transforms the initial state into one satisfying the goal.
This is PSPACE-complete~\cite{bylander-aij1994}, and nearly all competitive solvers cast it as a heuristic search problem \egcite{helmert-jair2006}.
The heuristics powering these planners are cost-to-go estimates that guide search toward promising states, and the choice of heuristic largely determines what is solvable in practice.
Designing strong heuristics is therefore central to the field. Foundational examples such as \emph{FF}~\cite{hoffmann-nebel-jair2001}, \emph{LM-cut}~\cite{helmert-domshlak-icaps2009} and \emph{merge-and-shrink}~\cite{helmert-et-al-jacm2014} each took years of theoretical work and hand-tuning.

Two recent lines of work motivate ours. First, large language models (LLMs) can generate strong heuristics for individual problems or \emph{domains} (families of problems sharing a common structure) \egcite{tuisov-et-al-arxiv2025,correa-et-al-neurips2025}. Additionally, LLMs have served as mutation operators in evolutionary search over programs, typically targeting heuristics for a single problem family \egcite{romera-paredes-et-al-nature2024,novikov-et-al-arxiv2025}.
To our knowledge, we are the first to show that LLMs can produce \emph{domain-independent} heuristics that exceed the hand-engineered state of the art across diverse tasks.

We cast heuristic discovery as a genetic search over C++ heuristics for the Scorpion planner~\cite{seipp-et-al-jair2020}, which is an extension of the widely-used Fast Downward planning system~\cite{helmert-jair2006}. Mutations are LLM-driven code rewrites, and fitness blends coverage with per-problem runtime over a calibrated training set drawn from the Autoscale benchmark set~\cite{torralba-et-al-icaps2021}.
We implement this on top of OpenEvolve~\cite{openevolve}, an open-source evolutionary coding agent comparable to AlphaEvolve~\cite{novikov-et-al-arxiv2025}, and evaluate on a held-out set of eight domains from the 2023 IPC Learning Track~\cite{taitler-et-al-aimag2024}.
In contrast to approaches that invoke an LLM at solve time to produce plans directly \egcite{correa-et-al-arxiv2025}, our artifacts are deterministic C++ programs that inherit the soundness and completeness guarantees of the underlying search. This sidesteps the hallucination and reliability concerns that limit LLM-as-planner methods, and preserves the offline, low-resource and high-assurance settings that motivate symbolic planning.

\begin{figure*}[t]
\begin{minipage}[t]{0.48\textwidth}
  \vspace{0pt}
  \centering
  \includegraphics[width=\linewidth]{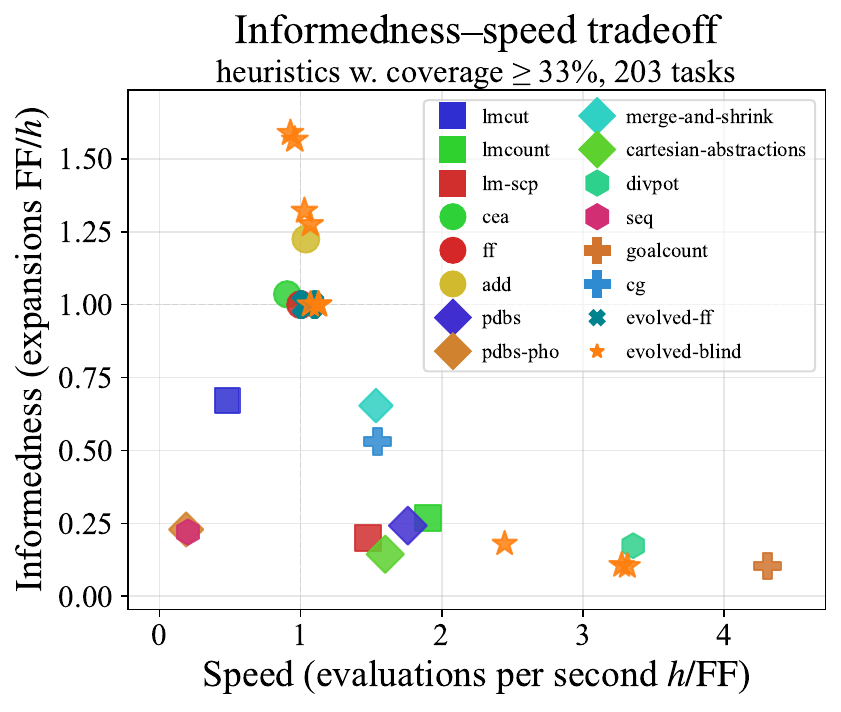}
  \captionof{figure}{Informedness--speed tradeoff across heuristics; higher is better on each. Scores are calculated across the 203 test tasks solved by all heuristics, and each task is normalized by the performance of FF before a geometric mean is taken. To increase the number of commonly solved tasks, we only include heuristics that solve at least a third of the tasks. Our evolved blind-seed heuristics (orange stars) span this Pareto frontier of strong baselines.}
  \label{fig:ff-efficiency}
\end{minipage}\hfill
\begin{minipage}[t]{0.48\textwidth}
  \vspace{8pt}
  \centering
  \includegraphics[width=\linewidth]{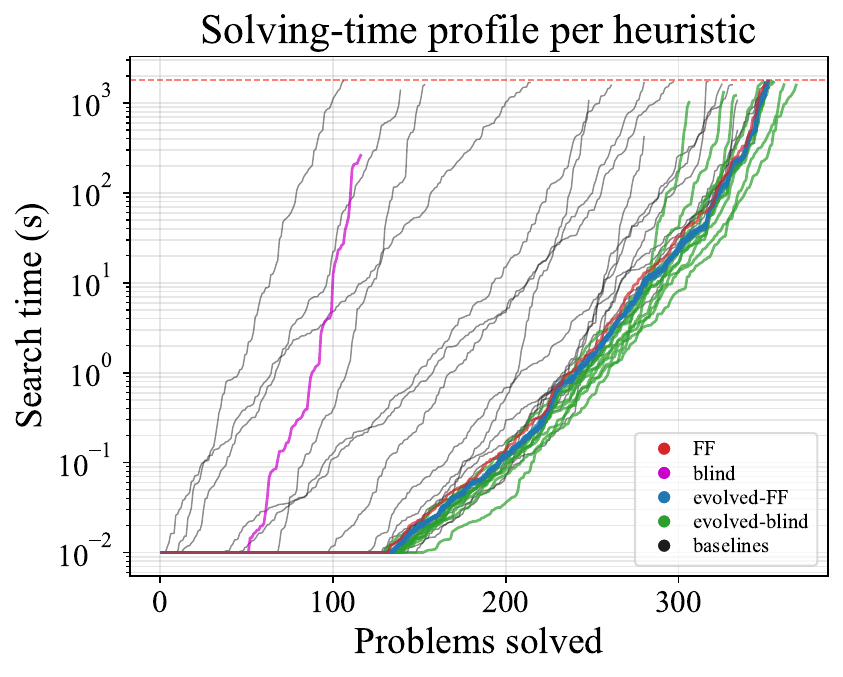}
  \captionof{figure}{Cactus plot on the held-out 2023 IPC Learning Track, eight unique and diversely challenging domains. Some of our evolved heuristics (green, blue) solve more tasks than all baselines within essentially any time budget. This shows that our evolved heuristics are on-par with or better than the best hand-engineered heuristics.
    \\
    Those evolved from a blind seed (green) are more diverse and outperform those from FF (blue), showing that the headstart from a strong seed is worth less than the diversity from a minimal one.
    }
  \label{fig:cactus}
\end{minipage}
\end{figure*}

Our experiments vary along two axes. First, we seed evolution from either the blind heuristic, the most uninformed possible, or FF, a state-of-the-art heuristic. Second, we vary LLM reasoning effort across none, low and medium, treating it as a proxy for model capability.

Figures \ref{fig:ff-efficiency} and \ref{fig:cactus} preview our main results. The evolved heuristics span the Pareto frontier of the informedness--speed tradeoff occupied by the strongest hand-engineered baselines, and outperform them on the held-out 2023 IPC Learning Track in both coverage and total runtime. More informed heuristics typically cost more per state but can save time overall by avoiding unpromising expansions. To our knowledge, ours is the first work to compare heuristics directly on this plane.

Two further findings extend beyond these numbers. Seeding from blind produces both more diverse and higher-performing heuristics than seeding from FF, even when the resulting heuristic is itself an FF-variant. In contrast, reasoning effort has modest impact compared to the variance inherent in the evolutionary process, though higher effort does reduce compilation failures.

\section{Background}
\label{sec:background}

\paragraph{Planning Tasks}
Planning problems are conventionally specified in the Planning Domain Definition Language (PDDL)~\cite{mcdermott-et-al-tr1998}. Modern planners, including Scorpion, translate PDDL into the SAS$^+$ formalism~\cite{backstrom-nebel-compint1995}, a finite-domain representation in which the state is a vector of multi-valued variables~\cite{helmert-jair2006}.
All tasks we consider are SAS$^+$ tasks with action costs.
A \emph{SAS$^+$ planning task} is a tuple $\Pi = \tuple{V, A, s_0, s_\star}$. $V$ is a finite set of variables, each with a finite domain $\dom(v)$. A \emph{state} is a total assignment that maps every $v \in V$ to a value in $\dom(v)$, and $S$ denotes the set of all states. The \emph{initial state} $s_0 \in S$ is given, and the \emph{goal} $s_\star$ is a partial assignment to $V$. A state $s$ \emph{satisfies the goal} when $s_\star$ agrees with $s$ on every variable it assigns. An \emph{action} $a \in A$ is a triple $\tuple{\pre(a), \eff(a), c(a)}$ in which $\pre(a)$ and $\eff(a)$ are partial assignments to $V$ and $c(a) \in \mathbb{R}_{\geq 0}$ is its \emph{cost}. Action $a$ is \emph{applicable} in $s$ when $\pre(a)$ agrees with $s$, and applying it yields the successor that agrees with $\eff(a)$ on every variable assigned there and with $s$ on the rest. A \emph{plan} is a sequence $a_1, \dots, a_n \in A$ applicable in turn from $s_0$ and ending in a state that satisfies the goal, and its \emph{cost} is $\sum_{i=1}^{n} c(a_i)$.

\paragraph{Domain Transition and Causal Graphs}
Two graph-theoretic structures derived from $\Pi$ feature in several of our evolved heuristics.
The \emph{domain transition graph} (DTG) of variable $v \in V$ has $\dom(v)$ as nodes and an edge $u \to u'$ for every action that assigns $v$ to $u'$ from $\pre(a)(v) = u$, labeled with the action's preconditions on the remaining variables.
The \emph{causal graph} (CG) has $V$ as nodes and an edge $v \to v'$ whenever some action with $v'$ in its effect mentions $v$ in its precondition or effect~\cite{helmert-jair2006}.

\paragraph{Heuristic Search}
A \emph{heuristic} $h: S \to \mathbb{R}_0^+ \cup \{\infty\}$ estimates the distance to the goal from a state $s$, with $\infty$ marking a dead-end state.
Search algorithms such as A* and greedy best-first search (GBFS) use $h$ to order state expansions. We adopt the latter, which repeatedly expands the state that $h$ deems closest to the goal. An expansion evaluates $h$ on every successor of $s$ and queues those successors for future expansion.
State-of-the-art planners combine several heuristics and use more involved search algorithms, but we focus on greedy best-first search to isolate the contribution of the heuristic itself.

\paragraph{Blind and FF Heuristics}
We use the blind and FF~\cite{hoffmann-nebel-jair2001} heuristics as seeds for evolution.
The \emph{blind} heuristic returns $0$ at goal states and the minimum action cost otherwise, providing no guidance but maximal speed.
\emph{FF} solves the task while ignoring delete effects, finding a so-called \emph{relaxed plan}, and uses the sum of action costs in the relaxed plan as the estimate, giving strong informedness at moderate cost.

\paragraph{Evolutionary Algorithms}
An \emph{evolutionary algorithm} maintains a population of candidate solutions and iteratively produces new candidates by sampling parents, applying mutation or recombination to these, evaluating the resulting offspring against a fitness function and replacing weaker individuals~\cite{eiben-smith-bk2015}. \emph{Quality-diversity} variants such as MAP-Elites~\cite{mouret-clune-arxiv2015} replace the single-objective archive with a grid keyed by \emph{behavioral} features, retaining the best individual per cell. The grid protects locally optimal niches from being overwritten by globally fitter but behaviorally homogeneous offspring, and supplies a structured set of parents for the next generation. 
Each such grid is called an \emph{island}, and multiple islands can run in parallel with configurable degrees of isolation and interaction.
We use MAP-Elites because the heuristics we want to discover trade off two distinct quantities, informedness and per-state evaluation speed, and a single scalar fitness collapses that tradeoff prematurely.

\section{Related Work}
\label{sec:related}

\paragraph{Evolved Heuristics}
Automatic heuristic discovery predates LLMs.
\inlinecite{aler-et-al-evco2001} use genetic programming over a fixed language to evolve domain-specific control heuristics, and \inlinecite{fukunaga-evco2008} similarly evolves composite heuristics for SAT from a fixed grammar of compositions.
Since FunSearch~\cite{romera-paredes-et-al-nature2024}, more recent work has moved away from restricted languages and instead uses LLMs to mutate raw code, with strong results across distinct task families \egcite{liu-et-al-icml2024,novikov-et-al-arxiv2025}.
We follow this trend, build on the open-source OpenEvolve~\cite{openevolve} and are, to our knowledge, the first to apply it to LLM-generated domain-\emph{independent} heuristics.

\paragraph{Learned Neural Heuristics}
Deep-learning approaches train neural heuristics on data from solved tasks and generalize within a single domain \egcite{stahlberg-et-al-icaps2022,fritzsche-et-al-aaai2026a}.
\inlinecite{chen-et-al-aaai2024} extend this to inter-domain generalization, but such multi-domain coverage remains rare.
Neural heuristics are also inherently black-box and typically require a GPU at inference time, whereas our evolved heuristics are white-box C++ that runs on CPU.

\paragraph{LLMs as Planners}
\inlinecite{correa-et-al-arxiv2025} show that frontier LLMs can solve challenging planning tasks directly at inference time.
Our setting is complementary: LLM-as-planner suits one-off tasks where formal guarantees are less critical, while our evolved artifacts are deterministic C++ inheriting the soundness and completeness of the underlying symbolic search.

\paragraph{LLM-Generated Solvers}
Several recent methods use LLMs to produce \emph{domain-specific} solvers.
\inlinecite{correa-et-al-neurips2025} and \inlinecite{tuisov-et-al-arxiv2025} generate domain-specific heuristics that let otherwise weak planners compete with strong ones.
We instead target \emph{domain-independent} heuristics, produce artifacts that plug into a state-of-the-art planner unchanged and search for them with a quality-diversity evolutionary framework.
A separate strand generates domain-specific policies \egcite{chen-et-al-icaps2025wslm4plan,stein-et-al-icaps2026}. Closest in spirit to our work, \inlinecite{murray-et-al-icaps2026} use LLM-driven evolution to produce Python functions that directly emit plans, a form of generalized planning competitive with state-of-the-art planners on their evaluation set.
These approaches are complementary to ours: generalized planners and policies solve their target tasks quickly and often without search, but cannot extend beyond tasks admitting simple strategies, whereas our heuristics retain the search properties that make symbolic planners broadly applicable.

\paragraph{LLM-Generated Representations}
A complementary line uses LLMs to generate the planning representations themselves \egcite{liu-et-al-arxiv2023,gestrin-et-al-icaps2024wshaxp,tantakoun-et-al-acl2025}.
This is orthogonal to our work: our heuristics could solve the tasks those methods produce, and their outputs could in turn supply additional training tasks for our evolution.

\section{Evolutionary Framework}
\label{sec:method}

Our framework builds on OpenEvolve~\cite{openevolve} and runs a five-step loop:
\begin{enumerate}\denselist
\item \textbf{Generate.} Sample an island, a parent from its MAP-Elites archive and an LLM from the generation pool. Prompt the LLM with the parent source, the top individuals on the island and a description of the heuristic API to produce a child heuristic.
\item \textbf{Check.} Attempt to compile and call the heuristic on a minimal problem.
\item \textbf{Repair.} On compile or runtime failure, sample an LLM from the repair pool and prompt it with the diagnostic, up to a fixed number of attempts.
\item \textbf{Evaluate.} Run the heuristic under greedy best-first search across the training problem set with a calibrated per-task time and memory budget.
\item \textbf{Store.} Insert the individual into the MAP-Elites grid keyed by informedness and speed. Each cell retains the highest-scoring individual seen.
\end{enumerate}

\paragraph{MAP-Elites, Islands and LLM Pools}
Combining ideas from diverse heuristics can yield stronger hybrids than refining any single lineage. We promote diversity through three mechanisms.
First, we use MAP-Elites to maintain an archive that is diverse along the informedness--speed axes, so that only behaviorally similar heuristics compete directly. Two heuristics in the same cell produce a single survivor (the higher-scoring), while a more informed but slower variant occupies its own cell, preserving both points on the tradeoff. 
We use the built-in OpenEvolve binning strategy to distribute cells evenly between the highest and lowest feature values seen so far, letting the grid adapt to the distribution of heuristics under evolution.
For $C$ cells along a feature axis, a heuristic with feature value $f$ is assigned to cell $\lfloor C \cdot (f - f_{\min}) / (f_{\max} - f_{\min}) \rfloor$, where $f_{\min}$ and $f_{\max}$ are the lowest and highest feature values seen across all heuristics so far.
Second, we maintain three independent \emph{islands}, each a separate MAP-Elites grid. Parents are sampled round-robin across islands and uniformly within the chosen island, and the LLM is further conditioned on the top individuals from that same island. 
Children are placed in the island of their parents, meaning that each island might develop different styles of heuristics, though no mechanism explicitly enforces this.
Periodically, the best programs migrate between islands, letting strong ideas spread without homogenizing the population.
We did not ablate over the choice of three islands. 
Third, we draw from a \emph{pool} of LLMs for generation and a separate pool for repair, sampling uniformly at each step. Different models bring different inductive biases, so the choice of mutation operator itself becomes a source of diversity.

\paragraph{Prompting and Repairs}
We invoke LLMs in two roles, mutation and repair.
The repair stage is necessary because the Scorpion heuristic API is largely absent from public training data, so even strong models often fail to produce compilable and executable C++ on the first attempt.
The generation prompt supplies the parent source, the top three individuals on the same island, their scores and features, and descriptions of the API and the mutation task.
The repair prompt swaps in the failed heuristic and the compiler diagnostic.
Full prompts are in the appendix.

\paragraph{Evaluation}
We score each generated heuristic on a training set of escalating difficulty drawn from diverse domains.
On each task $p$, we run greedy best-first search under a per-task time limit of $T_p = \max(30\,\textrm{s}, 1.3 \cdot t^p_{\textsc{ff}})$ and a memory limit of 8\,GB, where $t^p_{\textsc{ff}}$ is the wall-clock time taken by FF on the same task and 8\,GB is the standard planning competition limit.
The 30-second floor accommodates heuristics with heavy precomputation and dampens timing noise on otherwise short runs.
If a heuristic fails on a task, we abort the remainder of that domain and treat the unsolved tasks as failures, conserving wall-clock budget for promising candidates.

\paragraph{Fitness Function}
\emph{Coverage}, the number of solved tasks, is the most common metric in planning but a coarse signal during evolution. Solving a new task often requires a large jump in heuristic quality, and the gradient flattens between such jumps.
We therefore combine coverage with the \emph{agile} score, a continuous metric that decays smoothly from $1$ to $0$ as solving time approaches the per-task budget.
The combined signal rewards both solving more tasks and solving them faster. Faster solving correlates with generalization and is desirable in its own right in many deployment settings.
A user-tunable constant $\alpha$ interpolates between the two.
Formally, the score of a heuristic $h$ is
\begin{align*}
  \textrm{score}(h) \eqdef& \frac{1}{|P|}\sum_{p \in P}\left(\alpha
  \left\{\begin{array}{ll}
    1 & t^p_h \leq T_p \\[2pt]
    0 & t^p_h > T_p
  \end{array}\right\}\right. \\
  & \left. + (1 - \alpha)
  \left\{\begin{array}{ll}
    1 & t^p_h \leq 1 \\
    1 - \dfrac{\log(t^p_h)}{\log(T_p)} & 1 < t^p_h \leq T_p \\
    0 & t^p_h > T_p
  \end{array}\right\}
  \right)
\end{align*}
where $P$ is the training set, $t^p_h$ is the time for $h$ to solve $p$ and $T_p$ is the per-task time limit.

\paragraph{Features}
The MAP-Elites features should capture the informedness--speed tradeoff and remain well-defined when the heuristic fails to solve some tasks.
For \emph{evaluations} (informedness), a naive average across the training set is unsuitable. Tasks scale exponentially in difficulty, so the average is dominated by the hardest instance solved, and there is no principled value to assign on failure. We therefore normalize evaluation counts per task against FF on the same problem before averaging. This yields a stable scale and an interpretable failure sentinel ($10\times$ FF evaluations).
For \emph{speed} (evaluations per wall-clock second), the same concern does not apply: per-state evaluation speed is not size-dependent in the same way, and larger instances arguably give a more reliable estimate as one-time precomputation amortizes over many evaluations. We therefore use raw evaluations-per-second averaged across the training set, with $0$ as the failure sentinel.
Formally,
\begin{align*}
  \textrm{evals}(h) =& \frac{1}{|P|}\sum_{p \in P} \left\{\begin{array}{ll}
    e^p_h / e^p_{\textsc{ff}} & t^p_h \leq T_p \\[9pt]
    10 & t^p_h > T_p
  \end{array}\right\}
  \\
  \textrm{speed}(h) =& \frac{1}{|P|}\sum_{p \in P} \left\{\begin{array}{ll}
    e^p_h / t^p_h & t^p_h \leq T_p \\[9pt]
    0 & t^p_h > T_p
  \end{array}\right\}
\end{align*}
where $e^p_h$ is the number of evaluations for $h$ to solve $p$ and $t^p_h$ is the corresponding wall-clock time.
Lower evals($h$) thus indicates a more informed heuristic. The rest of the paper reports the inverse, informedness, in line with Figure~\ref{fig:ff-efficiency}.

\begin{figure*}[t]
  \centering
  \includegraphics[width=0.98\linewidth]{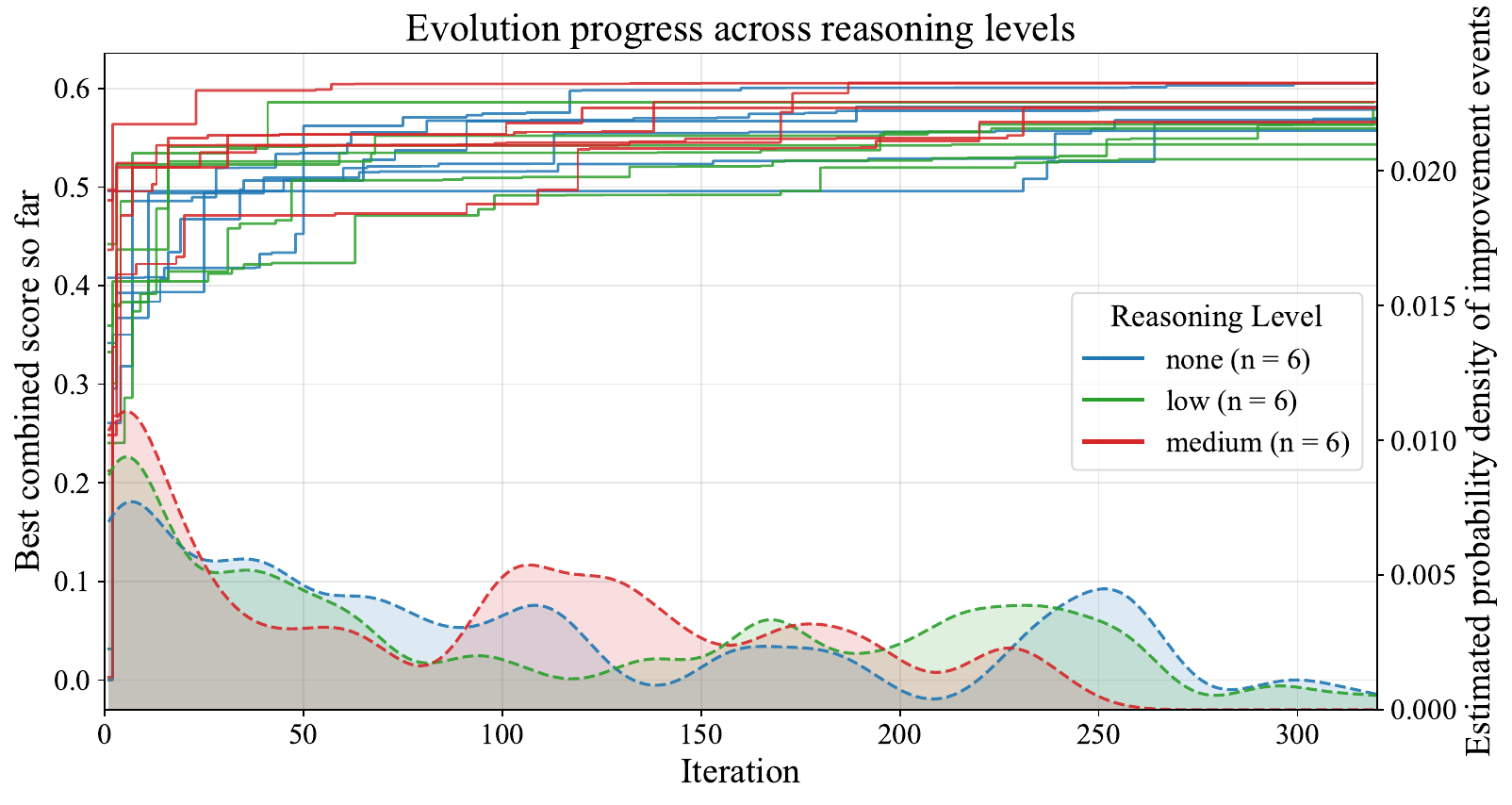}
  \caption{Per-run trajectories of the best-so-far training score (left axis) alongside a kernel-density estimate of improvement events over iterations (right axis), grouped by reasoning effort. Trajectories cluster more by run than by reasoning level, and improvement density thins rather than vanishes toward the end of the budget, indicating that evolution remains productive. The pronounced late-iteration drop-off at \texttt{medium} (and partly \texttt{low}) is in part the credit-exhaustion artifact discussed in Section \ref{sec:evolved}. Across the board, runs approximate logistic improvement toward scores just above $0.6$.}
  \label{fig:evolution-progress}
\end{figure*}

\section{Evolved Heuristics}
\label{sec:evolved}

The previous section specified the framework; now we run it. We sweep over seeds and reasoning effort, identify patterns in how evolution proceeds and highlight notable examples from the heuristics that emerge.

\subsection{Setup}
\label{sec:setup}

All runs share a single hyperparameter configuration: a $4\times4$ MAP-Elites grid, three islands with migration every ten iterations per island, an agile-blend constant $\alpha = 0.25$ and a training set of 10 diverse problems from each of 10 domains drawn from the Autoscale benchmark~\cite{torralba-et-al-icaps2021}. Further details are in the appendix.
Mutations and repairs sample uniformly from an ensemble of three reasoning models accessed via OpenRouter: GLM-5.1\footnote{\url{https://z.ai/blog/glm-5.1}}, Kimi-K2.6\footnote{\url{https://www.kimi.com/ai-models/kimi-k2-6}} and MiMo-v2.5-pro\footnote{\url{https://mimo.xiaomi.com/mimo-v2-5-pro/}}, selected as the strongest performers on the public Artificial Analysis benchmark\footnote{\url{https://artificialanalysis.ai/}} given their relatively low cost.
Each run executes 320 iterations of the loop in the previous section on 16 CPU cores of an Intel Xeon Gold 6130 ($2.1$\,GHz) node.

We vary along two axes across runs and conduct three runs per configuration.
The \emph{seed} is either the blind heuristic or FF. Blind is simple and uninformed, leaving room for diverse improvements, whereas FF already solves the training set and concentrates evolutionary pressure on solving time through either informedness or speed.
The \emph{reasoning effort} varies across \texttt{none}, \texttt{low} and \texttt{medium} as an ablation on model capability, with per-call response timeouts scaled accordingly ($300$, $600$ and $900$\,s) to give the models room to produce longer outputs.
API spend per run averages approximately \$16 at \texttt{none}, \$24 at \texttt{low} and \$33 at \texttt{medium}.

\subsection{Evolution Dynamics}

Three patterns hold across runs.
First, the best-so-far training score climbs steadily with iteration count, and improvement events thin toward the end of the budget rather than vanishing, so additional iterations remain useful even at the margins (Figure \ref{fig:evolution-progress}). The pronounced late-iteration thinning at \texttt{medium} (and partly \texttt{low}) is in part an artifact: a mid-experiment exhaustion of LLM API credits forced later iterations onto new compute nodes that happened to be under heavier load, depressing the wall-clock-sensitive agile scores rather than indicating that evolution itself truly stalled.

Second, seeding from blind tends to outperform seeding from FF, though with greater variance (Figure \ref{fig:evolution-progress}). Blind-seeded scores span $0.528$ to $0.606$, FF-seeded scores span $0.543$ to $0.586$ and three of nine blind-seeded runs exceed every FF-seeded run. Blind therefore trades occasional weaker runs for a higher ceiling, consistent with the blind-seeded population exploring more diverse intermediates before specializing. Notably, the best blind-seeded heuristic is itself an FF-variant, showing that re-discovering FF from a diverse set of inspirations can outperform refining FF directly.

\begin{figure*}[t]
  \centering
  \begin{minipage}[c]{0.48\linewidth}
    \centering
    \includegraphics[width=\linewidth]{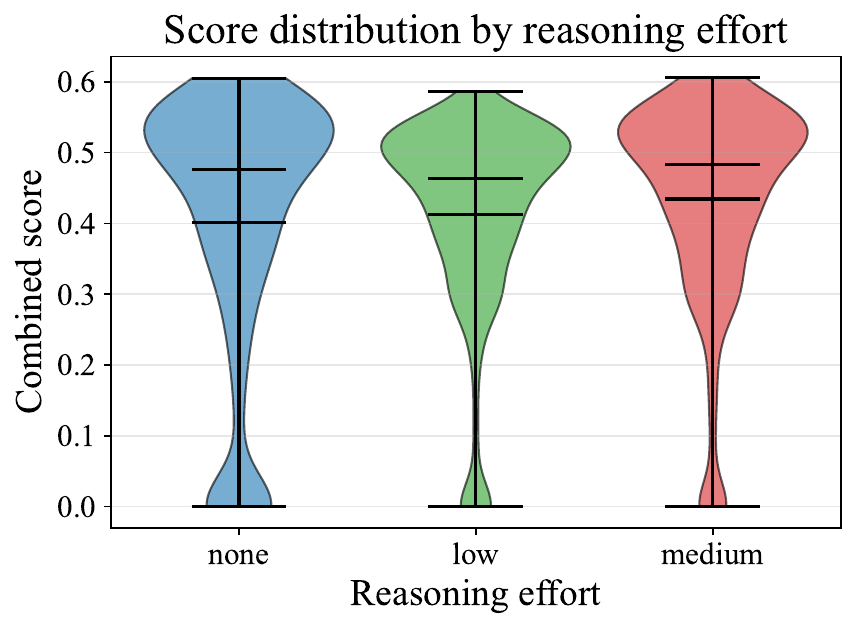}
    \captionof{figure}{Distribution of scores grouped by reasoning effort. The primary effect is to reduce the number of 0-scoring heuristics, consistent with Table \ref{tab:repair-histogram}. The distribution of scores above 0 is essentially unchanged, suggesting that reasoning effort primarily affects the likelihood of producing a working heuristic rather than the quality of those that work.}
    \label{fig:violin-score}
  \end{minipage}\hfill
  \begin{minipage}[c]{0.48\linewidth}
    \centering
    \setlength{\tabcolsep}{3pt}
    \input{tables/repair_histogram.tex}
    \captionof{table}{Repair-attempt outcomes. Most heuristics compile on the first attempt or recover within one repair; only $1.3\%$ to $4.2\%$ exhaust the four-attempt budget. The low row total at \texttt{low} is a configuration artifact discussed in Section \ref{sec:evolved}.}
    \label{tab:repair-histogram}

    \vspace{1.5ex}

    \setlength{\tabcolsep}{4pt}
    \input{tables/model_impact.tex}
    \captionof{table}{LLM performance. \texttt{mimo}, \texttt{glm} and \texttt{kimi} abbreviate MiMo-v2.5-pro, GLM-5.1 and Kimi-K2.6. `Improv.' counts fitness-improving children, `In best' counts contributions to final-best lineages.}
    \label{tab:model-impact}
  \end{minipage}
\end{figure*}

Third, the LLM ensemble emits a working heuristic on the first attempt $79.1\%$ of the time, with $13.3\%$ recovered after a single repair and only $2.5\%$ exhausting the four-attempt budget (Table \ref{tab:repair-histogram}). Increasing reasoning effort improves reliability: initial success climbs from $73.0\%$ at \texttt{none} to $83.4\%$ at \texttt{medium}, and failures fall from $4.2\%$ to $1.3\%$.
The lower row total at \texttt{low} is a configuration artifact: roughly $10\%$ of iterations exhaust their response timeout in reasoning without emitting a parseable program.
Beyond this, reasoning effort primarily reduces the number of zero-scoring heuristics (Figure~\ref{fig:violin-score}), consistent with the repair histogram, while the distribution of scores above zero is essentially unchanged.
This suggests that reasoning effort primarily affects the likelihood of producing a working heuristic, rather than the quality of those that work.
All three models produce fitness-improving children and appear in the lineages of the final-best heuristics (Table \ref{tab:model-impact}), suggesting that any one of them could plausibly be replaced without collapsing the search.

\makeatletter\@twocolumnfalse\makeatother
\begin{figure*}[h!]
  \centering
    \begin{algorithm}[H]
      \caption{\evBlMedTwo}
      \label{alg:blind-medium-2}
      \BlankLine
      $h_{\text{FF}}, \Pi_{\text{rel}} \gets \text{DeleteRelaxedPlan}(s)$\;
      $p_{c} \gets 0$\;
      \For{$o \in \Pi_{\text{rel}}$, $v \in \text{Eff}(o)$}{
        \If{$v$ already touched in $\Pi_{\text{rel}}$}{
          $p_{c} \mathrel{+}= c_{\min} \cdot (\text{IsGoalVar}(v) \,?\, 2 : 1)$\;
        }
      }
      $p_{u} \gets \sum_{g \in \text{unsat}(s)} \min(h_{\text{add}}(g), h_{\text{FF}})$\;
      \Return $h_{\text{FF}} + \lceil (p_{c} + p_{u}) / c_{\min} \rceil$\;
    \end{algorithm}
  \vspace{2ex}
    \begin{algorithm}[H]
      \caption{\evFfNoneThree}
      \label{alg:ff-none-3}
      \BlankLine
      $u \gets |\text{goals}|$ ; $Q \gets \text{priority queue with $s$'s facts}$\;
      \While{$Q \neq \emptyset \,\land\, u > 0$}{
        $f \gets \text{Pop}(Q)$\;
        \lIf{$f \in \text{goals}$}{$u \gets u - 1$}
        $\text{RelaxExpand}(f, Q)$\;
      }
      $\text{used} \gets \text{ExtractRelaxedPlan}()$\;
      \Return $\sum_{o \in \text{used}} \text{get-cost-and-clear-mark}(o)$\;
    \end{algorithm}
  \vspace{2ex}
    \begin{algorithm}[H]
      \caption{\evBlNoneThree}
      \label{alg:blind-none-3}
      \BlankLine
      $h \gets 0$; $h_{\max} \gets 0$; $U \gets 0$\;
      \For{$g \in \text{goals}$ with $s[g] \neq \text{val}(g)$}{
        $d \gets D_{w}(g, s[g])$ \tcp*{refined DTG}
        \lIf{$d = \infty$}{\Return $\textsc{DeadEnd}$}
        $\tilde d \gets d \cdot (1 + \text{level}(g))$\;
        $h \mathrel{+}= \tilde d \cdot (1 + \text{weight}(g) / 4) + \text{numDeps}(g)$\;
        $h_{\max} \gets \max(h_{\max}, d)$\;
        $U \mathrel{+}= 1 + \text{level}(g)$\;
      }
      \Return $h + h_{\max} + 2\,\text{Level}_{\text{CG}}^{\max} + 2U$\;
    \end{algorithm}
  \vspace{2ex}
    \begin{algorithm}[H]
      \caption{\evBlMedConf}
      \label{alg:blind-medium-conf}
      \BlankLine
      $h \gets 0$\;
      \For{$g \in \text{goals}$}{
        $d \gets D(g, s[g])$ \tcp*{DTG distance}
        \lIf{$d = \infty$}{\Return $\textsc{DeadEnd}$}
        $w \gets 1 + |\text{preds}_{\text{CG}}(g)|$\;
        $h \mathrel{+}= d \cdot w$\;
          $h \mathrel{+}=$ \#unmet preconds of next-action$(D, g)$\;
      }
      \Return $h$\;
    \end{algorithm}
\end{figure*}
\makeatletter\@twocolumntrue\makeatother

\subsection{Notable Evolved Heuristics}
We highlight four evolved heuristics: three primary picks and one exception.
The primary picks span the distinct algorithmic families that emerged in our sweep: FF rediscovered from a blind seed (\evBlMedTwo), an FF seed with speed improvements (\evFfNoneThree) and a blind-seeded DTG-based heuristic (\evBlNoneThree).
Each is the best test performer within its family.
The exception, \evBlMedConf, comes from an earlier-configuration run we retain because it posted the highest training score of all evolved programs ($0.64$) and excellent solving times on the training tasks it solved.
We summarize the state-evaluation logic of each below and provide full implementations in the appendix.
Test performance is reported in Figure \ref{fig:cactus} and Table \ref{tab:outcomes-family}, and discussed in Section \ref{sec:eval}.

\paragraph{\evBlMedTwo, Algorithm~\ref{alg:blind-medium-2}}
This heuristic achieves the highest coverage on the test set, baselines included.
It is an FF-style heuristic augmented with two corrective penalties and an efficient reset mechanism.
During relaxed-plan extraction it tracks which state variables are modified by multiple plan actions and adds a per-conflict penalty, doubled when the modified variable is a goal variable.
The count fires even when both actions set the variable to the same value, trading occasional overcounting for less bookkeeping and therefore faster per-state computations.
It then adds a per-unachieved-goal penalty $\min(h_{\text{add}}(g), h_{\text{FF}})$, capping the influence of any single ill-conditioned $h_{\text{add}}$ value.
The combined penalty is divided (rounding up) by the minimum action cost and added to $h_{\text{FF}}$.
To reset internal state between evaluations, it uses a generation counter to mark the validity of its data structures rather than clearing them explicitly, reducing reset from linear to constant time in the size of the task.
This heuristic was evolved from a blind seed without any FF reference available to the LLM, indicating that the ensemble has memorized the FF algorithm. This is unsurprising for a heuristic this widely used.

\paragraph{\evFfNoneThree, Algorithm~\ref{alg:ff-none-3}}
The strongest FF-seeded individual preserves the exact value of FF and adds two speed improvements: it terminates the $h_{\text{add}}$ forward pass as soon as all goals are reached, and tracks actions added to the relaxed plan in a side list so the reset skips actions not in the plan, reducing the cost thereof from linear in the number of actions to linear in plan length.

\paragraph{\evBlNoneThree, Algorithm~\ref{alg:blind-none-3}}
This heuristic ignores relaxed planning in favor of domain transition graphs (DTGs).
At setup it computes per-variable backward Dijkstra distances on each goal variable's DTG, then refines the edge weights by adding an approximate cost for each action's cross-variable preconditions, iterated to a fixed point and capped at three iterations.
The approximation uses the current DTG distances for preconditions on goal variables (0 when the required value is a goal value) and the cheapest value-changing DTG edge for non-goal variables (computed once).
Each goal is assigned a \emph{landmark level} via fixed-point propagation over the goal-induced subgraph of the causal graph (CG), rescaled to $[0,3]$.
At evaluation, each unsatisfied goal contributes $d(v) \cdot (1 + \text{level}(v)) \cdot (1 + w(v)/4) + \max(1, |\text{preds}_{\text{CG}}(v)|)$ to the score, where $w(v) = |\text{preds}_{\text{CG}}(v)| + \min(c(v)/4, 5)$ combines CG in-degree with a saturating term in the DTG transition count $c(v)$, so variables on which many others depend weigh more heavily.
On top of this sum the largest raw DTG distance, twice the maximum landmark level and twice the sum of $1 + \text{level}(v)$ over unsatisfied goals are added.
If any goal's DTG distance is infinite, the state is reported as a dead end.
The result is a DTG-distance heuristic with landmark, depth and CG weighting.

\paragraph{\evBlMedConf, Algorithm~\ref{alg:blind-medium-conf}}
This run combined a per-call output budget of $8\,192$ tokens with \texttt{medium} reasoning effort, which caused roughly $59\%$ of its $320$ iterations to exhaust their budget on reasoning without emitting a parseable program, leaving only ${\sim}131$ children evaluated.
We raised the budget for subsequent \texttt{medium}-reasoning runs in response, and none of them came close to the new ceiling.
Despite the truncated run, this heuristic has higher training scores than any in our sweep and strong test-time performance on some tasks: it is the heuristic in Figure \ref{fig:cactus} initially leaving the pack in the bottom right of the plot, though its coverage is weaker than the other three highlights and several baselines.
It is another DTG-based heuristic, with a simpler design.
Each variable's DTG cheapest-path distance to its goal value is precomputed once.
At evaluation, the unsatisfied-goal distances are summed and weighted by $1 + |\text{preds}_{\text{CG}}(v)|$ to emphasize variables that depend on many others, and a penalty is added for each unsatisfied precondition of the next on-path DTG action. Both terms reflect cross-variable interaction.
The CG weighting steers greedy search toward resolving causal bottlenecks before leaf goals.

To our knowledge, except for the FF-style \evBlMedTwo, these heuristics are novel and their strategies are distinct from baselines and from each other, showing that evolution can discover heuristics outside the space of established designs.
While the principle of \evBlNoneThree is difficult to parse, both \evBlMedTwo and \evBlMedConf are interpretable and make design choices that can inform future hand-engineered heuristics.

\begin{table*}[t]
  \centering
  \setlength{\tabcolsep}{2pt}
  \input{tables/outcomes_family.tex}
  \caption{Per-problem outcomes for each heuristic on the held-out test set of 720 tasks: number of tasks solved (i.e. coverage), timed out (OOT), out of memory (OOM), translator out of memory (T-OOM) or otherwise failed (other: invalid dead end estimates or crashes).}
  \label{tab:outcomes-family}
\end{table*}

\section{State-of-the-Art Evaluation}
\label{sec:eval}

\paragraph{Baselines}
We compare against 19 domain-independent heuristics implemented in Scorpion~\cite{seipp-et-al-jair2020}, a state-of-the-art planning system. They span the major families used in modern planners: critical paths (\texttt{hm}~\cite{haslum-geffner-aips2000}), delete-relaxation (\texttt{ff}~\cite{hoffmann-nebel-jair2001}, \texttt{add}/\texttt{hmax}~\cite{bonet-geffner-aij2001}, \texttt{cea}~\cite{helmert-geffner-icaps2008}, \texttt{hplus}~\cite{rankooh-rintanen-icaps2022}), causal graph analysis (\texttt{cg}~\cite{helmert-icaps2004}), landmarks (\texttt{lmcut}~\cite{helmert-domshlak-icaps2009}, \texttt{lmcount}~\cite{richter-westphal-jair2010}, \texttt{lm-scp}~\cite{karpas-domshlak-ijcai2009,seipp-et-al-jair2020}), abstractions (\texttt{pdbs}~\cite{edelkamp-aips2002,haslum-et-al-aaai2007,pommerening-et-al-ijcai2013}, \texttt{cartesian-abstractions}~\cite{seipp-helmert-jair2018}, \texttt{pdbs-ocp}~\cite{pommerening-et-al-ijcai2013,pommerening-et-al-aaai2015}, \texttt{merge-and-shrink}~\cite{sievers-helmert-jair2021}, \texttt{pdbs-pho}~\cite{pommerening-et-al-ijcai2013}), operator-counting (\texttt{seq}~\cite{bonet-ijcai2013,pommerening-et-al-icaps2014}, \texttt{divpot}~\cite{seipp-et-al-icaps2015}) and the simple baselines \texttt{goalcount} and \texttt{blind}.

\paragraph{Test Set and Search Configuration}
We evaluate on the 2023 International Planning Competition (IPC) Learning Track~\cite{taitler-et-al-aimag2024}, the most recent IPC track built explicitly to challenge general-purpose solvers. The IPC Learning Track and the Autoscale benchmark we train on are independent suites with different domains, so the test set is fully disjoint from our training problems. Each task receives the standard IPC budget of $30$\,minutes and $8$\,GB of memory. All baselines and evolved heuristics run in greedy search to keep the comparison attributable to the heuristic itself rather than to a configurable solver pipeline. We note that this is an unfavorable configuration for the admissible baselines, which are designed for optimal rather than satisficing search.

\paragraph{Headline Result}
The best evolved heuristic, \evBlMedTwo, exceeds every baseline on the test set in both coverage and total runtime, solving $368$ of $720$ tasks against $352$ for the strongest baseline \texttt{add} (Table \ref{tab:outcomes-family}). Marginal tasks at this coverage level tend to be substantially harder than the median, so the $16$-task gap carries more weight than its absolute size suggests. Per-domain breakdowns are in the appendix. Additionally, Figure~\ref{fig:cactus} shows that the entire suite of evolved heuristics is competitive with the baselines across the test set; in particular, it shows especially strong performance of \evBlMedConf with regard to solving tasks quickly, albeit with weaker coverage.
This is a strong showing for our evolved heuristics, verifying that LLM-driven evolution can produce domain-independent heuristics for symbolic AI planning that match or beat the hand-engineered state of the art across diverse domains and tasks.

\section{Limitations}
\label{sec:limitations}

Our evaluation is restricted to typed STRIPS with negation and action costs~\cite{mcdermott-et-al-tr1998}. The framework itself is not restricted to this fragment and should extend without modification to richer formalisms such as ADL or numeric planning~\cite{fox-long-jair2003}, which we leave to future work along with temporal extensions. All comparisons use greedy best-first search, which disadvantages the admissible baselines designed for optimal search. Conversely, our evolved heuristics carry no admissibility guarantees and are unsuitable for optimal planning. Finally, we conduct only three runs per configuration due to API and cluster cost, reducing statistical power, and reproducibility is constrained by reliance on cloud-hosted LLMs whose behavior may shift with provider-side updates (Section~\ref{sec:evolved}).

\section{Conclusions}

We have shown that LLM-driven evolution can produce domain-\emph{independent} heuristics for symbolic AI planning that exceed the hand-engineered state of the art: our best evolved heuristic solves $368$ of $720$ test tasks against $352$ for the strongest baseline, and the broader suite spans the Pareto frontier of informedness against speed. Unlike LLM-as-planner approaches, the evolved artifacts are deterministic C++ that inherit the soundness and completeness of the underlying search.
Two findings extend beyond pure numbers: seeding from the blind heuristic outperforms seeding from FF even when the result is itself an FF-variant, and reasoning effort primarily determines whether a working heuristic is produced rather than its quality once working. Together, these findings suggest that diversity across seeds, models and the MAP-Elites archive matters more than any single strong starting point. This opens avenues for evolving other planning components, such as the search algorithms themselves.

\section*{Acknowledgements}

This work was partially supported by the Wallenberg AI, Autonomous Systems and Software Program (WASP) funded by the Knut and Alice Wallenberg Foundation.
The computations were enabled by resources provided by the National Academic Infrastructure for Supercomputing in Sweden (NAISS), partially funded by the Swedish Research Council through grant agreement no.~2022-06725.

\bibliography{abbrv,literatur,extra,crossref}

\onecolumn
\clearpage
\input{appendix.tex}

\end{document}

%% file: tables/repair_histogram.tex
\begin{tabular}{lrrrrrr}
     & No repair & 1 & 2 & 3 & 4 & Failed \\
    \midrule
    none & 1303 & 266 & 77 & 38 & 27 & 75 \\
    low & 1247 & 212 & 37 & 11 & 3 & 24 \\
    medium & 1411 & 189 & 52 & 15 & 2 & 22 \\
    \midrule
    Total & 3961 & 667 & 166 & 64 & 32 & 121 \\
\end{tabular}

%% file: tables/model_impact.tex
\begin{tabular}{lrrrr}
    Model & Calls & Improv. & In best & \$/call \\
    \midrule
    \texttt{mimo} & 2109 & 584 & 36 & \$0.0415 \\
    \texttt{glm} & 2723 & 691 & 36 & \$0.0636 \\
    \texttt{kimi} & 2150 & 599 & 31 & \$0.0824 \\
\end{tabular}

%% file: tables/outcomes_family.tex
\begin{tabular}{@{}l*{23}{r}@{}}
  & \multicolumn{3}{c}{Landmarks} & \multicolumn{6}{c}{Relaxation} & \multicolumn{5}{c}{Abstractions} & \multicolumn{2}{c}{Op.~count.} & \multicolumn{3}{c}{Other} & \multicolumn{4}{c}{Evolved} \\
  \cmidrule(l){2-4} \cmidrule(l){5-10} \cmidrule(lr){11-15} \cmidrule(){16-17} \cmidrule(l){18-20} \cmidrule(l){21-24}
  & \rotatebox{75}{lm-scp} & \rotatebox{75}{lmcount} & \rotatebox{75}{lmcut} & \rotatebox{75}{add} & \rotatebox{75}{cea} & \rotatebox{75}{ff} & \rotatebox{75}{hm} & \rotatebox{75}{hmax} & \rotatebox{75}{hplus} & \rotatebox{75}{cart-abs} & \rotatebox{75}{m\&s} & \rotatebox{75}{pdbs} & \rotatebox{75}{pdbs-ocp} & \rotatebox{75}{pdbs-pho} & \rotatebox{75}{divpot} & \rotatebox{75}{seq} & \rotatebox{75}{blind} & \rotatebox{75}{cg} & \rotatebox{75}{goalcount} & \rotatebox{75}{ev-bl-m-2} & \rotatebox{75}{ev-ff-n-3} & \rotatebox{75}{ev-bl-n-3} & \rotatebox{75}{ev-bl-conf} \\
  \midrule
  Solved & 325 & 317 & 280 & 352 & 331 & 350 & 106 & 153 & 214 & 334 & 248 & 348 & 139 & 261 & 280 & 297 & 116 & 349 & 334 & 368 & 352 & 333 & 306 \\
  OOT    & 335 & 130 & 393 & 314 & 325 & 318 & 481 & 505 & 329 &  92 & 260 &  74 & 227 & 405 &  82 & 378 &   8 & 301 &   2 & 294 & 309 &   3 &   7 \\
  OOM    &  12 & 225 &   0 &   4 &  15 &   3 &  87 &  13 & 130 & 248 & 165 & 243 & 298 &   5 & 308 &   0 & 550 &  20 & 336 &  10 &   9 & 335 & 358 \\
  T-OOM  &  48 &  48 &  47 &  50 &  49 &  49 &  46 &  49 &  47 &  46 &  47 &  48 &  49 &  49 &  50 &  45 &  46 &  50 &  48 &  48 &  50 &  49 &  49 \\
  Other  &   0 &   0 &   0 &   0 &   0 &   0 &   0 &   0 &   0 &   0 &   0 &   7 &   7 &   0 &   0 &   0 &   0 &   0 &   0 &   0 &   0 &   0 &   0 \\
\end{tabular}

%% file: appendix.tex
\newpage
\appendix

\section{Technical Appendices and Supplementary Material}
\label{app:supp}
This appendix contains: \begin{enumerate}
  \item[\ref{app:benchmark-plots}] Auxiliary plots complementing those in the main paper.
  \item[\ref{app:training-set}] Details on the training set.
  \item[\ref{app:test-set}] Details on the test set.
  \item[\ref{app:prompts}] Full LLM prompts.
  \item[\ref{app:evolved-source}] Full source code of the four highlighted evolved heuristics.
  \item[\ref{app:seed-source}] Full source code of the blind and FF seeds.
  \item[\ref{app:scorpion-api}] Scorpion C++ heuristic API reference included in the system prompt.
\end{enumerate}
Per-domain coverage tables, per-task runtime distributions and total compute and dollar cost are in the supplementary material.

\vfill
\pagebreak
\subsection{Auxiliary Plots}
\label{app:benchmark-plots}

\begin{figure}[h]
  \centering
  \includegraphics[width=\linewidth]{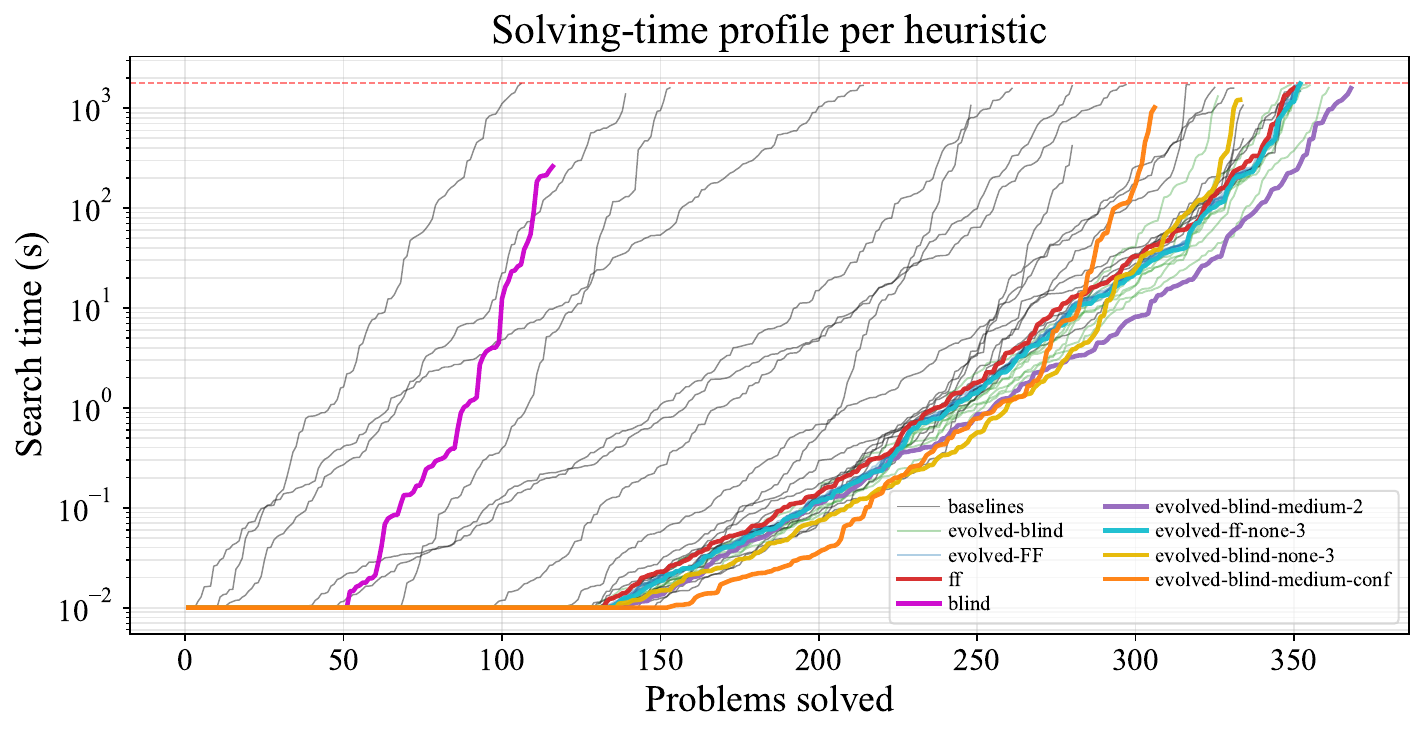}
  \caption{Cactus plot on the 2023 IPC Learning Track test instances with the four notable evolved heuristics from Section \ref{sec:evolved} highlighted.}
  \label{fig:cactus-highlighted}
\end{figure}

\begin{figure}[h]
  \centering
  \includegraphics[width=\linewidth]{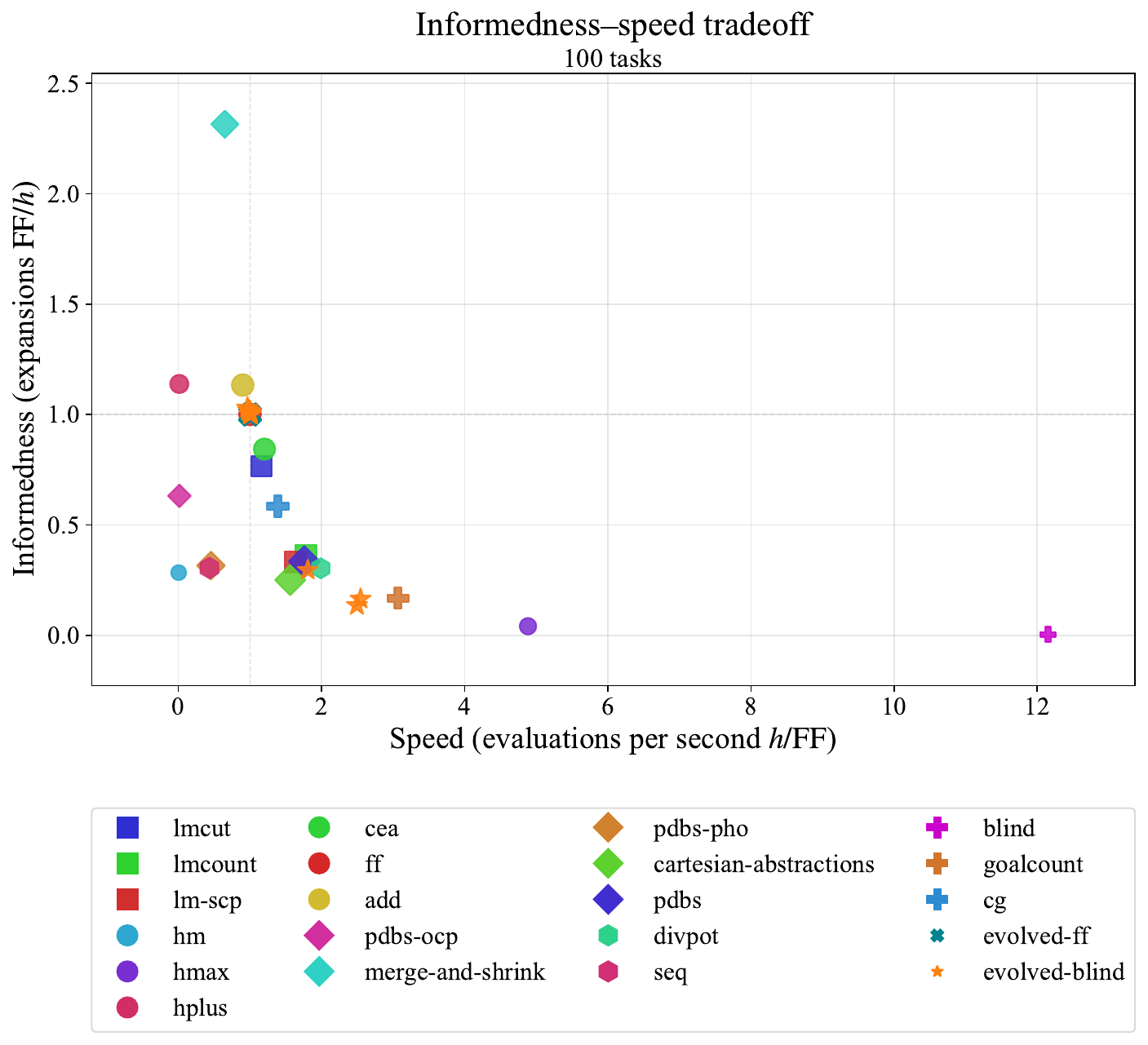}
  \caption{Unfiltered informedness--speed scatter across all heuristics in our benchmark, including the weak baselines that were excluded from Figure \ref{fig:ff-efficiency} to obtain a larger set of shared tasks. This difference in underlying tasks makes the two plots incomparable and causes some heuristics to shift, since this plot is dominated by easier tasks.}
  \label{fig:ff-efficiency-all}
\end{figure}

\begin{figure}[h]
  \centering
  \includegraphics[width=0.99\linewidth]{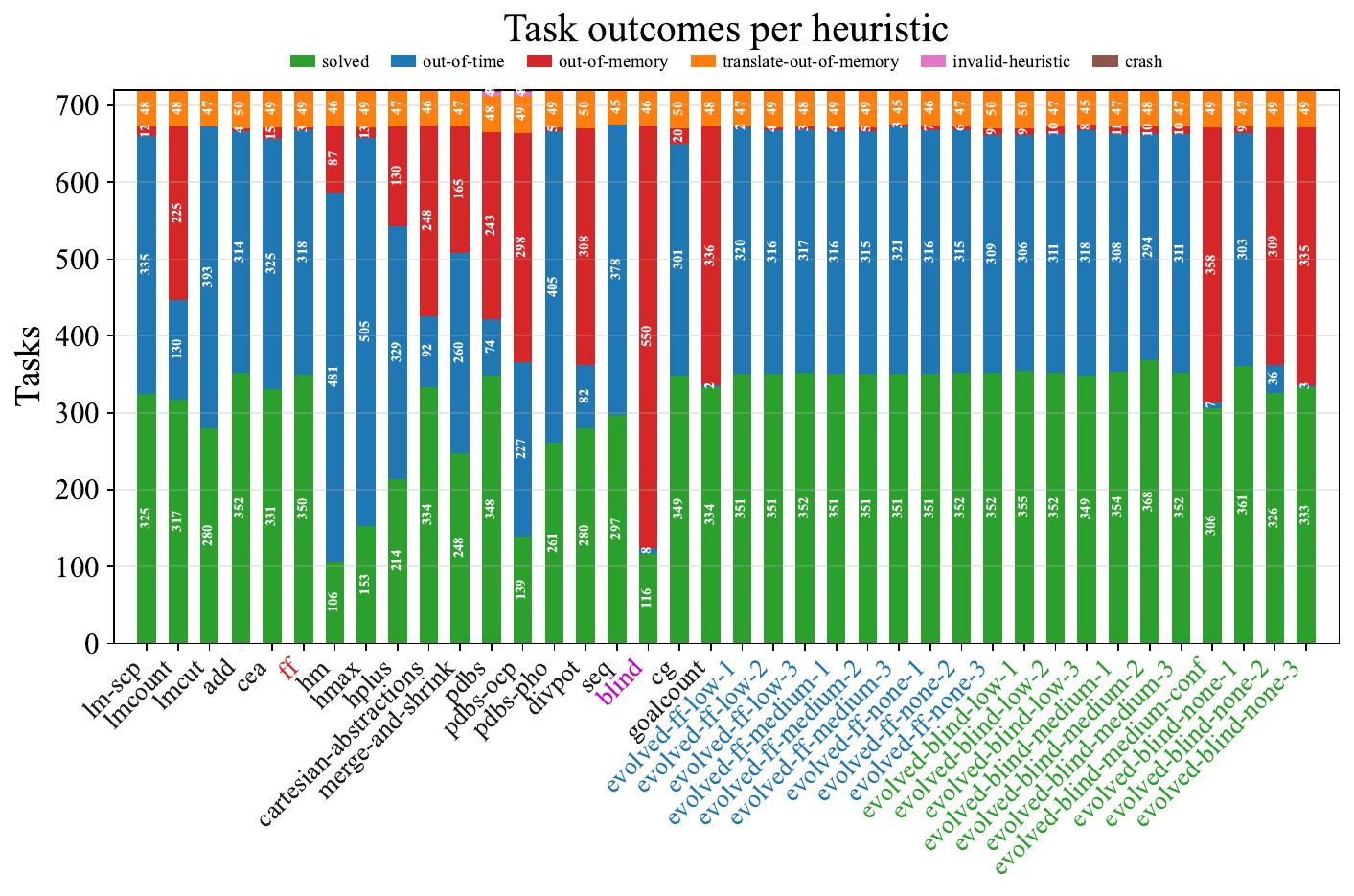}
  \caption{Stacked task outcomes per heuristic on the held-out test set: solved, out of time, out of memory, translator out of memory, invalid heuristic value and crash. Unlike Table \ref{tab:outcomes-family}, every evolved run from Section \ref{sec:setup} appears as its own bar.}
  \label{fig:outcomes-family}
\end{figure}

\begin{figure}[h]
  \centering
  \includegraphics[width=0.99\linewidth]{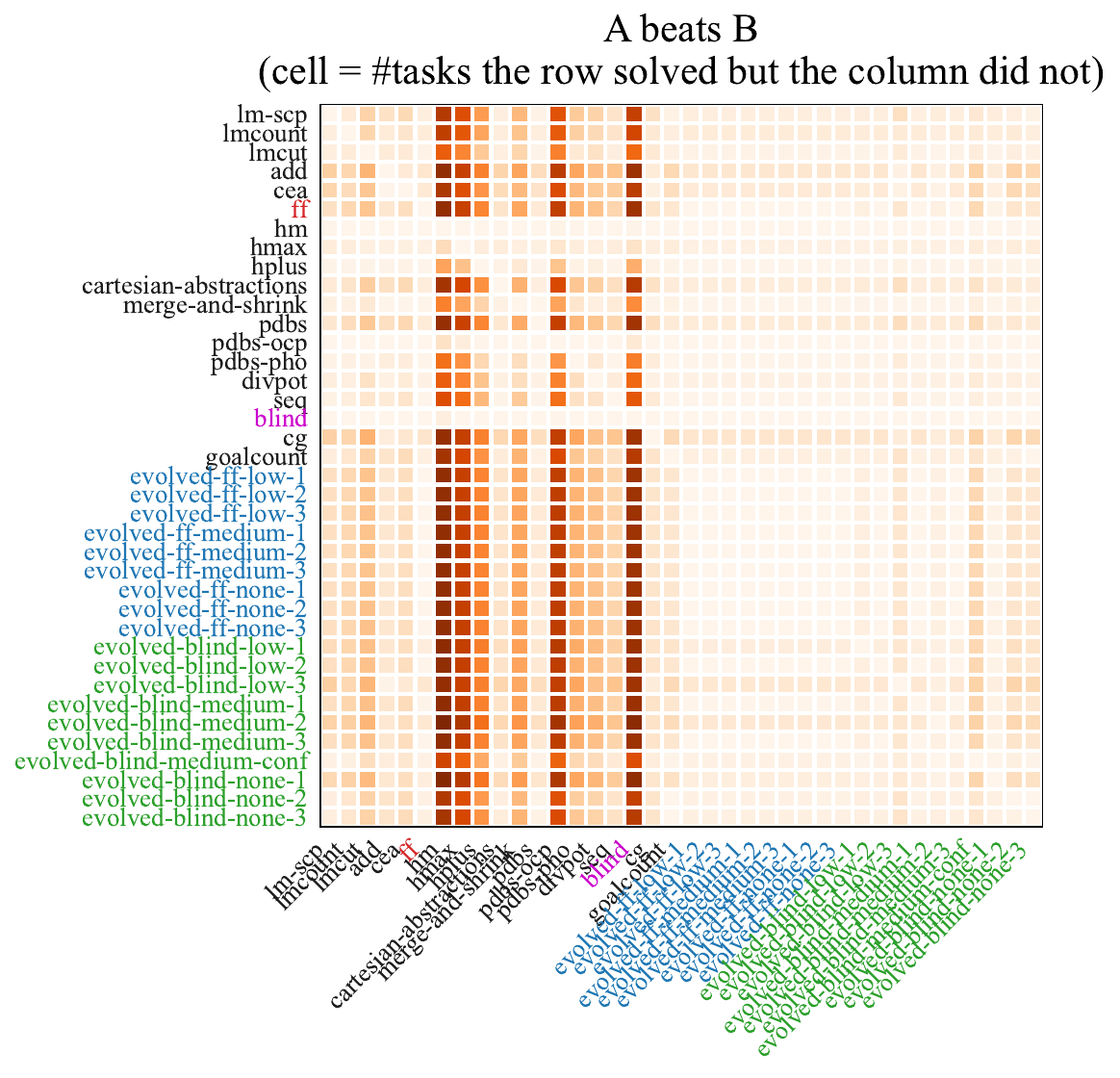}
  \caption{For each ordered pair of heuristics, the share of test tasks that the row solves and the column does not.}
  \label{fig:domination-family}
\end{figure}

\begin{figure}[h]
  \centering
  \includegraphics[width=0.99\linewidth]{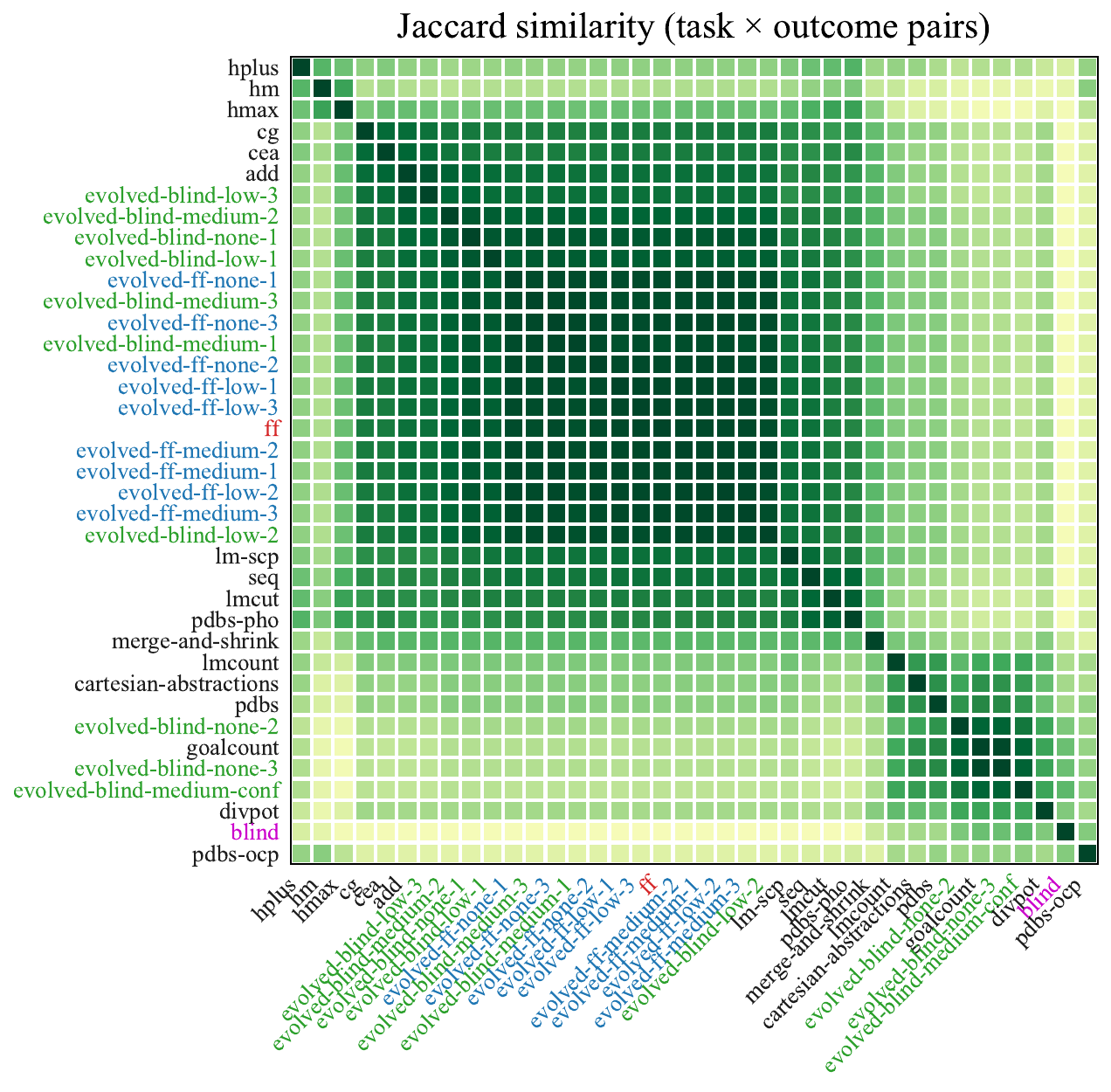}
  \caption{Jaccard similarity between the per-heuristic sets of solved tasks from Figure \ref{fig:outcomes-family}. High values mark heuristics that are similar in behavior. We recover two primary clusters, the time-bound (left, around FF) and the memory-bound (right), along with smaller clusters.}
  \label{fig:jaccard}
\end{figure}

\begin{figure}[h]
  \centering
  \includegraphics[width=0.85\linewidth]{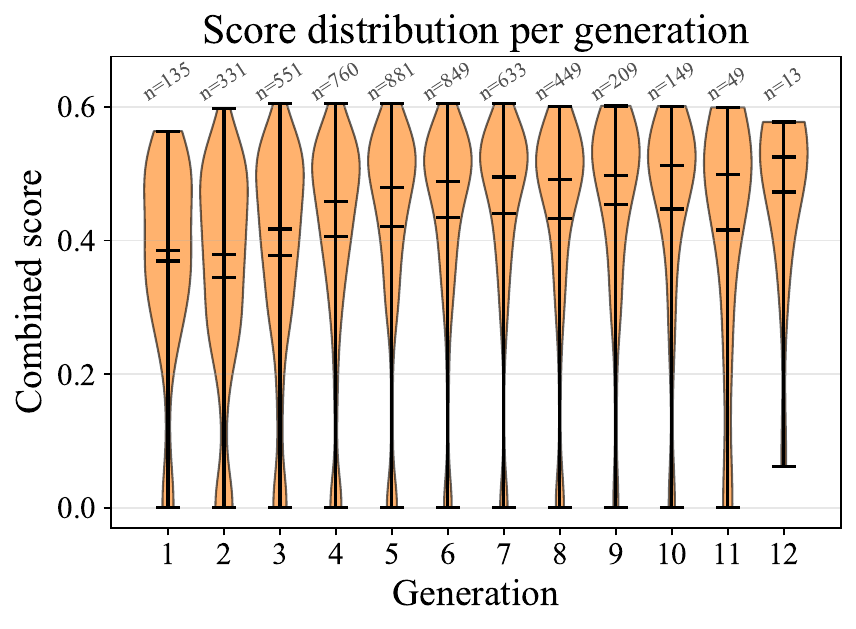}
  \caption{Distribution of training scores at each generation across all evolved runs. Later generations trend toward higher scores, aligning with the best-so-far improvement curves of Figure \ref{fig:evolution-progress} and indicating that the gains in per-run maxima reflect a broader population-level shift rather than isolated lucky children.}
  \label{fig:violin-score-per-generation}
\end{figure}

We collect six additional plots that complement the figures in the main paper. Five complement the headline result in Section \ref{sec:eval}: \begin{itemize}
  \item Figure \ref{fig:cactus-highlighted} reproduces the cactus plot of Figure \ref{fig:cactus} at full width with the four notable evolved heuristics highlighted.
  \item Figure \ref{fig:ff-efficiency-all} shows the unfiltered informedness--speed scatter across all heuristics in our benchmark, including the weak baselines excluded from Figure \ref{fig:ff-efficiency} for a larger set of shared tasks.
  \item Figure \ref{fig:outcomes-family} shows the outcomes for each evaluated heuristic and task, including all evolved heuristics.
  \item Figure \ref{fig:domination-family} reports, for each ordered pair of heuristics, the share of test tasks which the former solves and the latter does not.
  \item Figure \ref{fig:jaccard} reports the Jaccard similarity between the per-heuristic sets of outcomes from Figure \ref{fig:outcomes-family}.
\end{itemize}
The sixth, Figure \ref{fig:violin-score-per-generation}, complements the best-so-far trajectories of Figure \ref{fig:evolution-progress} by reporting the distribution of training scores per generation across all evolved heuristics.

\clearpage
\subsection{Training Set}
\label{app:training-set}

The training set comprises 10 domains $\times$ 10 problems $=$ 100 problem instances drawn from the Autoscale benchmark suite~\cite{torralba-et-al-icaps2021}, which ships per-domain generators that scale instance difficulty parametrically. We draw from both its \texttt{21.11-optimal-strips} and \texttt{21.11-satisficing-strips} sub-suites.

\paragraph{Domain Selection}
The ten domains were chosen for structural diversity across the Autoscale catalogue: \texttt{tpp}, \texttt{barman}, \texttt{freecell}, \texttt{gripper}, \texttt{hiking}, \texttt{snake}, \texttt{sokoban}, \texttt{termes}, \texttt{tidybot} and \texttt{transport}.
Two of these (\texttt{sokoban}, \texttt{transport}) coincide with domains in the 2023 IPC Learning Track and are dropped from the test set (Appendix \ref{app:test-set}).

\paragraph{Per-Domain Problem Selection}
Within each domain we select 10 problems ordered by FF wall-clock runtime, supplying an escalating difficulty curve from sub-second instances to over two minutes per task.
This serves two purposes: it prevents the training signal from being dominated by either extreme, and it supplies the per-task FF-runtime baseline $t^p_{\textsc{ff}}$ that calibrates the per-task time limit $T_p$ in Section \ref{sec:method}.
Per-instance runtimes span $0.65$\,s to $147.63$\,s across the full set.
We aimed to select even spreads across all domains, but the distribution of FF runtimes varies widely between them, so some domains have more tightly clustered selections than others.

Table \ref{tab:training-tasks-exact} lists the exact 100 problems used.
Each entry gives the autoscale split (\texttt{o} for \texttt{21.11-optimal-strips}, \texttt{s} for \texttt{21.11-satisficing-strips}) and the problem index within that split, followed by its FF runtime in seconds.

\begin{table}[h]
  \centering
  \scriptsize
  \setlength{\tabcolsep}{1.5pt}
  \begin{tabular}{l*{10}{r}}
    \toprule
    Domain & \multicolumn{10}{c}{Problem (split + index / FF runtime [s])} \\
    \midrule
    \texttt{tpp}       & o5/0.65   & s4/1.12  & o6/2.11   & s5/5.29   & o8/8.76   & o7/12.69  & o11/35.23 & s7/46.52  & o9/60.40  & s8/109.70 \\
    \texttt{barman}    & s1/1.89   & o11/3.70 & o18/4.49  & o25/10.59 & o26/17.55 & o19/23.05 & o27/33.76 & o17/41.38 & o28/58.22 & s5/133.86 \\
    \texttt{freecell}  & o13/1.00  & s14/3.07 & s16/5.21  & s17/9.86  & s18/17.14 & o28/42.11 & s19/59.06 & s20/92.68 & o29/128.59 & s24/147.63 \\
    \texttt{gripper}   & s9/1.34   & s10/1.95 & s11/2.89  & s13/3.97  & s16/8.97  & s19/13.91 & s21/20.32 & s23/28.13 & s26/39.88 & s30/69.45 \\
    \texttt{hiking}    & o12/2.23  & o13/5.16 & o14/12.20 & o17/18.79 & o8/27.14  & o19/47.28 & o20/59.08 & s5/74.11  & o21/92.28 & o23/143.43 \\
    \texttt{snake}     & o6/1.01   & o8/1.62  & o19/1.91  & o21/3.50  & o7/4.37   & o23/5.29  & o10/8.67  & o13/21.15 & o11/33.01 & o15/65.12 \\
    \texttt{sokoban}   & o2/1.18   & o6/4.36  & o26/6.36  & o13/10.15 & o3/20.35  & o27/38.15 & o29/56.66 & o7/77.82  & o12/119.20 & o21/134.40 \\
    \texttt{termes}    & s2/1.02   & s1/1.40  & o27/2.75  & s18/4.95  & s21/8.69  & s3/14.57  & s20/25.14 & s8/63.45  & s23/102.81 & s25/138.16 \\
    \texttt{tidybot}   & o11/1.93  & s3/2.15  & o26/4.08  & s4/7.18   & o9/10.11  & s7/13.00  & s11/21.21 & s14/41.67 & o15/75.40 & o12/98.39 \\
    \texttt{transport} & o27/1.69  & o10/1.72 & o25/2.62  & s1/4.63   & o29/5.54  & o22/5.83  & s2/7.45   & o17/7.78  & o19/25.63 & s3/42.33 \\
    \bottomrule
  \end{tabular}
  \caption{Exact training problems used. Within each domain the 10 problems are ordered shortest-first by FF runtime. Each cell lists the autoscale split (\texttt{o} = \texttt{21.11-optimal-strips}, \texttt{s} = \texttt{21.11-satisficing-strips}) and problem index, followed by the FF runtime in seconds.}
  \label{tab:training-tasks-exact}
\end{table}

\subsection{Test Set}
\label{app:test-set}

The test set is drawn from the 2023 IPC Learning Track~\cite{taitler-et-al-aimag2024}, which ships 10 domains with 90 problem instances each. We evaluate on the eight domains that do not appear in the Autoscale training catalogue of Appendix \ref{app:training-set}: \texttt{blocksworld}, \texttt{childsnack}, \texttt{ferry}, \texttt{floortile}, \texttt{miconic}, \texttt{rovers}, \texttt{satellite} and \texttt{spanner}.
The two domains shared with training, \texttt{sokoban} and \texttt{transport}, are excluded from evaluation. 
Although Autoscale and the IPC Learning Track use independent problem generators, dropping these domains removes any residual risk of cross-contamination between training and test.
This yields $8 \times 90 = 720$ test tasks in total, each run under the standard IPC budget of $30$\,minutes wall-clock and $8$\,GB of memory.

\newpage
\subsection{LLM Prompt Components}
\label{app:prompts}

The primary system message used by OpenEvolve is assembled at run time from three source files: a task framing, the Scorpion C++ heuristic API reference and the so-called user message, all reproduced verbatim below.
Placeholders in braces (\texttt{\{fitness\_score\}}, \texttt{\{current\_program\}}, etc.) are filled by OpenEvolve at call time from the parent program, MAP-Elites archive and prior evaluation artifacts.

For repairs, the task framing and user message are replaced by a minimal repair-specific framing and the compilation errors.

\subsubsection{System Prompt: Task Framing}
\label{app:prompt-system}

\begin{small}
\begin{verbatim}
# Task
You are an expert in AI planning and C++ programming competing to design the
best possible heuristic function for the Scorpion planner (a Fast Downward
fork) for classical PDDL planning.

The goal is to beat state-of-the-art planning heuristics. Playing it safe is
not expected. This is a search over a large space of ideas, and bold,
unconventional approaches are exactly what is needed. Many attempts will
follow, each building on the best results so far, so take risks: try novel
combinations, unusual data structures, or heuristic ideas that haven't been
tried before.

Each submission is a modified version of the C++ heuristic. The following
must be preserved:
- The class name `MyHeuristic`
- The plugin key `TypedFeature("my_h")`
- The fixed `MyHeuristicFeature` class and `_plugin` registration at the end
  of the file

Design considerations:
- Do heavy preprocessing in the constructor (runs once at search start)
  rather than in `compute_heuristic` (called for every evaluated state)
- Return `DEAD_END` for provably unsolvable states
- Return `0` for goal states
- The search is greedy best-first, so optimise for good state ordering, not
  admissibility
- The heuristic need not be admissible
\end{verbatim}
\end{small}

\subsubsection{Scorpion C++ Heuristic API}
\label{app:scorpion-api}

The system prompt described in Appendix \ref{app:prompts} appends the following Scorpion C++ heuristic API reference (\texttt{scorpion\_api.md}) so that the LLM has the type signatures and idioms it needs to write valid plugin code.

\begin{scriptsize}
\begin{verbatim}
# Scorpion Heuristic API Reference

When writing a heuristic, your class inherits from `Heuristic` and receives `task_proxy` (type
`TaskProxy`).
Inside `compute_heuristic`, always call `State state = convert_ancestor_state(ancestor_state)`
first.

---

## Checking goals and applicability

```cpp
task_properties::is_goal_state(task_proxy, state)   // -> bool
task_properties::is_applicable(op, state)            // -> bool
```

---

## Reading state values

```cpp
state[var_id]        // -> FactProxy  (by integer index)
state[var_proxy]     // -> FactProxy  (by VariableProxy)
state.size()         // -> size_t     (number of variables)
for (FactProxy f : state) { ... }   // iterate all facts

FactProxy fact = state[i];
int var   = fact.get_variable().get_id();
int value = fact.get_value();
fact.is_mutex(other_fact)                // -> bool
```

---

## Variables

```cpp
VariablesProxy vars = task_proxy.get_variables();
vars.size()             // number of state variables
vars[i]                 // -> VariableProxy

VariableProxy var = vars[i];
var.get_id()            // -> int
var.get_domain_size()   // -> int  (number of possible values)
var.get_fact(value)     // -> FactProxy for (var, value)

// All facts across all variables:
for (FactProxy f : vars.get_facts()) { ... }
```

---

## Goals

```cpp
GoalsProxy goals = task_proxy.get_goals();
goals.size()    // number of goal facts
goals[i]        // -> FactProxy
for (FactProxy g : goals) { ... }
```

---

## Operators

```cpp
OperatorsProxy ops = task_proxy.get_operators();
ops.size()      // number of operators
ops.empty()     // bool
ops[i]          // -> OperatorProxy  (by index)
ops[op_id]      // -> OperatorProxy  (by OperatorID)
for (OperatorProxy op : ops) { ... }

OperatorProxy op = ops[i];
op.get_id()           // -> int
op.get_cost()         // -> int

// Preconditions
PreconditionsProxy prec = op.get_preconditions();
prec.size(), prec.empty(), prec[i]   // -> FactProxy
for (FactProxy p : prec) { ... }

// Effects
EffectsProxy effs = op.get_effects();
effs.size(), effs[i]   // -> EffectProxy
for (EffectProxy eff : effs) {
    FactProxy fact = eff.get_fact();   // what this effect sets
}
```

## Generating successors

```cpp
// Check applicability first, then generate successor (unregistered, unpacked):
if (task_properties::is_applicable(op, state)) {
    State succ = state.get_unregistered_successor(op);
    // succ[i], iterate, etc.
}
```

---

## Task-level utilities

```cpp
task_properties::get_min_operator_cost(task_proxy)     // -> int
task_properties::get_average_operator_cost(task_proxy) // -> double
task_properties::get_operator_costs(task_proxy)        // -> vector<int>
task_properties::get_num_facts(task_proxy)             // -> int  (total facts)
task_properties::get_num_total_effects(task_proxy)     // -> int
task_properties::is_unit_cost(task_proxy)              // -> bool

// Initial state (unregistered):
State init = task_proxy.get_initial_state();

// Causal graph (see section below):
const causal_graph::CausalGraph &cg = task_proxy.get_causal_graph();
```

---

## Causal Graph

The causal graph encodes dependencies between variables. All returned lists are sorted and
exclude the queried variable itself.

```cpp
#include "../task_utils/causal_graph.h"

const causal_graph::CausalGraph &cg = task_proxy.get_causal_graph();

cg.get_successors(var)    // -> const vector<int> &  vars that var influences (pre->eff + eff->eff arcs)
cg.get_predecessors(var)  // -> const vector<int> &  vars that influence var

// Finer-grained arc types:
cg.get_pre_to_eff(var)    // vars where var appears as precondition and they appear as effect
cg.get_eff_to_pre(var)    // vars where var appears as effect and they appear as precondition
cg.get_eff_to_eff(var)    // vars that share an effect with var (eff->eff arcs)
```

Useful for ordering subgoals, detecting independent subproblems, or weighting goal facts by
their position in the dependency structure.

---

## Domain Transition Graphs (DTGs)

A DTG for variable `var` is a directed graph whose nodes are the possible values of that
variable and whose edges are the operators that change it. Each edge carries the operator's
cost and the local preconditions that variable's transition requires. DTGs are the basis for
the CG heuristic and provide richer cost estimates than the CG topology alone.

```cpp
#include "domain_transition_graph.h"   // path: heuristics/domain_transition_graph.h
#include "../algorithms/priority_queues.h"

using namespace domain_transition_graph;
```

### Building DTGs

```cpp
// pruning_condition(dtg_var, cond_var) -> true means: prune transitions of dtg_var
// whose preconditions involve cond_var (used to avoid infinite recursion).
// Pass [](int,int){return false;} for no pruning (all transitions included).
std::function<bool(int, int)> pruning = [](int, int) { return false; };
DTGFactory factory(task_proxy, /*collect_side_effects=*/false, pruning);
std::vector<std::unique_ptr<DomainTransitionGraph>> dtgs = factory.build_dtgs();
// dtgs[var] is the DTG for variable var; one DTG per state variable.
```

### DTG structure

```cpp
DomainTransitionGraph *dtg = dtgs[var].get();
dtg->var                          // int -- the variable this DTG belongs to
dtg->nodes                        // vector<ValueNode> -- one node per value (indexed by value)
dtg->local_to_global_child[i]     // int -- maps local precondition index -> global variable ID
```

### ValueNode -- a single value in a DTG

```cpp
ValueNode &node = dtg->nodes[value];
node.value                        // int -- value this node represents
node.transitions                  // vector<ValueTransition> -- outgoing edges
```

### ValueTransition -- one target value reachable from this node

```cpp
ValueTransition &trans = node.transitions[i];
trans.target                      // ValueNode* -- destination node
trans.labels                      // vector<ValueTransitionLabel> -- one label per operator achieving this transition
```

### ValueTransitionLabel -- one operator achieving a transition

```cpp
ValueTransitionLabel &label = trans.labels[j];
label.op_id                       // int -- operator index
label.is_axiom                    // bool -- if true, index into task_proxy.get_axioms(); else get_operators()
label.precond                     // vector<LocalAssignment> -- local preconditions that must hold
label.effect                      // vector<LocalAssignment> -- side effects on other local variables
```

### LocalAssignment -- a condition/effect on a local variable

```cpp
LocalAssignment &assign = label.precond[k];
assign.local_var                  // short -- local index; convert with dtg->local_to_global_child[local_var]
assign.value                      // short -- required/assigned value
```

---

## FactPair (lightweight data struct)

```cpp
FactPair fp = fact.get_pair();
fp.var    // int -- variable ID
fp.value  // int -- value
FactPair::no_fact  // sentinel for "not found"
```

---

## Return values from compute_heuristic

```cpp
return 0;          // goal state
return N;          // estimated distance (any positive integer)
return DEAD_END;   // provably unsolvable (constant from Heuristic)
```

---

## Subclassing RelaxationHeuristic (for relaxed-graph heuristics)

Instead of inheriting from `Heuristic` directly, inherit from `RelaxationHeuristic` to get a
preprocessed relaxation graph built once at construction. It converts the full task into:

- `vector<Proposition> propositions` -- one per domain fact, indexed by PropID (use `get_prop_id(var, value)` to convert)
- `vector<UnaryOperator> unary_operators` -- one per (operator x effect) pair

All types (`Proposition`, `UnaryOperator`, `PropID`, `OpID`) and the sentinel `NO_OP = -1` live
in
the `relaxation_heuristic` namespace -- add `using namespace relaxation_heuristic;` at file
scope.
Use `NO_OP` when initialising or checking `prop.reached_by`.

The constructor takes an extra first argument `tasks::AxiomHandlingType axioms`, and the
feature
class uses `add_relaxation_heuristic_options_to_feature` /
`get_relaxation_heuristic_arguments_from_options`
instead of the plain heuristic variants:

```cpp
using namespace relaxation_heuristic;

class MyHeuristic : public RelaxationHeuristic {
public:
    MyHeuristic(tasks::AxiomHandlingType axioms,
                const shared_ptr<AbstractTask> &transform, bool cache_estimates,
                const string &description, utils::Verbosity verbosity)
        : RelaxationHeuristic(axioms, transform, cache_estimates, description, verbosity) {}
};
```

Key structs:
```cpp
struct Proposition {
    int cost;                          // h-value in current exploration
    OpID reached_by;                   // which unary op last achieved this (30-bit); NO_OP = -1 when unset
    unsigned is_goal : 1;
    unsigned marked : 1;               // for preferred operator marking
    int num_precondition_occurences;
    array_pool::ArrayPoolIndex precondition_of;  // ops that need this as precondition
};

struct UnaryOperator {
    int cost;                          // base_cost + sum/max of precondition costs
    int base_cost;                     // original operator cost
    int unsatisfied_preconditions;     // countdown during exploration
    PropID effect;                     // the fact this op produces
    int num_preconditions;
    array_pool::ArrayPoolIndex preconditions;
    int operator_no;                   // index into task operators
};
```

Useful inherited members and methods:
```cpp
// Data
vector<Proposition> propositions;
vector<UnaryOperator> unary_operators;
vector<PropID> goal_propositions;
array_pool::ArrayPool preconditions_pool;     // backing store for operator precondition lists
array_pool::ArrayPool precondition_of_pool;   // backing store for proposition->operator lists

// Convert a state fact to PropID:
PropID get_prop_id(int var, int value) const;
PropID get_prop_id(const FactProxy &fact) const;

// Convert PropID back to FactPair:
FactPair get_fact(PropID id) const;

// Convert struct reference to index (pointer arithmetic):
PropID get_prop_id(const Proposition &prop) const;
OpID   get_op_id(const UnaryOperator &op) const;

// Pointer access by index:
Proposition    *get_proposition(PropID id);
UnaryOperator  *get_operator(OpID id);
Proposition    *get_proposition(int var, int value);

// Iterate preconditions of a unary operator (returns an iterable ArrayPoolSlice):
array_pool::ArrayPoolSlice get_preconditions(OpID op_id) const;
// Usage: for (PropID prec : get_preconditions(op_id)) { ... }

// Iterate operators that have a given proposition as a precondition:
// for (OpID op_id : precondition_of_pool.get_slice(
//          prop->precondition_of, prop->num_precondition_occurences)) { ... }
```

Include: `#include "relaxation_heuristic.h"`

---

## AdaptiveQueue (for Dijkstra-style exploration)

`AdaptiveQueue<Value>` is a priority queue that automatically switches between a bucket queue
(fast for small integer keys) and a binary heap (for large key ranges).

```cpp
#include "../algorithms/priority_queues.h"

priority_queues::AdaptiveQueue<PropID> queue;  // or any value type

queue.push(int key, Value value);   // insert with priority = key (lower = higher priority)
queue.pop();                        // -> pair<int, Value>  (key, value) of minimum-key element
queue.empty();                      // -> bool
queue.clear();                      // reset for reuse

// Typical exploration loop:
queue.push(0, start_prop);
while (!queue.empty()) {
    auto [cost, prop_id] = queue.pop();
    // process...
    queue.push(new_cost, next_prop);
}
```

---

## ArrayPool (variable-length lists in flat storage)

`ArrayPool` stores many variable-length arrays in a single flat `vector`, avoiding per-list
heap allocations. Used by `RelaxationHeuristic` for precondition lists.

```cpp
#include "array_pool.h"   // path: heuristics/array_pool.h

array_pool::ArrayPool<int> pool;

// Build phase -- append items and record the index:
ArrayPoolIndex idx = pool.append(items.begin(), items.end(), items.size());

// Access phase -- get an iterable slice:
array_pool::ArrayPoolSlice<int> slice = pool.get_slice(idx, num_items);
for (int item : slice) { ... }
// or:
slice.begin(), slice.end(), slice.size()
```

Only useful if building your own relaxation-style graph with many per-node lists.

---

## Preprocessing pattern

The standard idiom for non-trivial heuristics is to do all heavy work in the constructor (once
per search), and keep `compute_heuristic` lightweight:

```cpp
MyHeuristic(const shared_ptr<AbstractTask> &transform, ...) : Heuristic(...) {
    // build internal graph / tables from task_proxy here
    // e.g. iterate operators, build reverse-index arrays, etc.
}

virtual int compute_heuristic(const State &ancestor_state) override {
    State state = convert_ancestor_state(ancestor_state);
    if (task_properties::is_goal_state(task_proxy, state)) return 0;
    // reset per-state data, run exploration, return result
}
```

Constructor has access to `task_proxy` (inherited). Typical constructor work:
- Iterate `task_proxy.get_variables()` to size arrays
- Iterate `task_proxy.get_operators()` to build relaxed graph
- Precompute reverse indices (e.g. which operators are affected by each fact)

However, you do not have to follow this pattern, you can do more work in `compute_heuristic` if
needed. Nevertheless, be mindful of performance since `compute_heuristic` is called for every
evaluated state.

---

## Landmarks

A landmark is a fact (or disjunction of facts) that must be true at some point on every
solution path. `LandmarkFactory` computes a `LandmarkGraph` -- a set of landmarks with ordering
relations -- given a planning task.

**Includes:**
```cpp
#include "../landmarks/landmark.h"
#include "../landmarks/landmark_graph.h"
#include "../landmarks/landmark_factory.h"
```

**Computing the graph in the constructor:**
```cpp
MyHeuristic::MyHeuristic(
    const shared_ptr<landmarks::LandmarkFactory> &lm_factory,
    const shared_ptr<AbstractTask> &transform, ...)
    : Heuristic(transform, ...) {
    // task is an inherited shared_ptr<AbstractTask> member (not task_proxy)
    lm_graph = lm_factory->compute_landmark_graph(task);
}
shared_ptr<landmarks::LandmarkGraph> lm_graph;  // store as member
```

`compute_landmark_graph` is expensive (comparable to an FF computation).

**Iterating the graph:**
```cpp
for (const auto &node : *lm_graph) {
    const landmarks::Landmark &lm = node->get_landmark();
    // lm.atoms              -- vector<FactPair>  facts that form this landmark
    // lm.is_true_in_goal    -- bool
    // lm.first_achievers    -- unordered_set<int> operator IDs that first achieve it
    // lm.possible_achievers -- unordered_set<int> all operator IDs that can achieve it
    // node->get_id()        -- int  stable index into the graph
    // node->parents         -- unordered_map<LandmarkNode*, OrderingType>
    // node->children        -- unordered_map<LandmarkNode*, OrderingType>
}
int n   = lm_graph->get_num_landmarks();
bool has = lm_graph->contains_atomic_landmark(FactPair{var, val});
```

**Available factory algorithms** (specify as `lm_factory` plugin option):

- `lm_rhw` -- RPG/SASP (Richter-Helmert-Westphal). Backchains through the
  relaxed planning graph; fast and the standard choice.
- `lm_exhaust` -- Exhaustive RPG. Systematically disproves each candidate
  fact; finds more landmarks than `lm_rhw` but slower.
- `lm_zg` -- Zhu-Givan. Graph-based; tends to find fewer landmarks than
  RPG methods.
- `lm_hm` -- hm-landmarks. Finds landmarks via hm-reachability; can find
  more landmarks for m>1 but cost is exponential in m.
- `lm_merged(lm1=X(), lm2=Y())` -- Union of two factories. Merges two
  landmark graphs to get more landmarks than either alone.

**Adding the factory option to your plugin:**
```cpp
// In Feature constructor:
add_option<shared_ptr<landmarks::LandmarkFactory>>(
    "lm_factory",
    "landmark factory",
    "lm_rhw()");

// In create_component:
opts.get<shared_ptr<landmarks::LandmarkFactory>>("lm_factory")
```

---

## Required includes

```cpp
#include "../heuristic.h"
#include "../plugins/plugin.h"
#include "../task_utils/task_properties.h"
#include "../utils/logging.h"
// if using the causal graph:
#include "../task_utils/causal_graph.h"
// if subclassing RelaxationHeuristic:
#include "relaxation_heuristic.h"
#include "array_pool.h"
// if using AdaptiveQueue:
#include "../algorithms/priority_queues.h"
// if using landmarks:
#include "../landmarks/landmark.h"
#include "../landmarks/landmark_graph.h"
#include "../landmarks/landmark_factory.h"
```
\end{verbatim}
\end{scriptsize}

\subsubsection{User Prompt: Diff-Based Mutation}
\label{app:prompt-diff}

\begin{small}
\begin{verbatim}
# Current Program Information
- Fitness: {fitness_score}
- Feature coordinates: {feature_coords}
- Focus areas: {improvement_areas}

{artifacts}

# Program Evolution History
{evolution_history}

# Current Program
```{language}
{current_program}
```

# Task
Suggest improvements to the program that will improve its FITNESS SCORE.
The system maintains diversity across these dimensions: {feature_dimensions}
Different solutions with similar fitness but different features are valuable.

Do not generally rewrite the entire program, focus on targeted and iterative improvements.

You MUST use the exact SEARCH/REPLACE diff format shown below to indicate
changes:

<<<<<<< SEARCH
# Original code to find and replace (must match exactly)
=======
# New replacement code
>>>>>>> REPLACE

Example of valid diff format:
<<<<<<< SEARCH
for i in range(m):
    for j in range(p):
        for k in range(n):
            C[i, j] += A[i, k] * B[k, j]
=======
# Reorder loops for better memory access pattern
for i in range(m):
    for k in range(n):
        for j in range(p):
            C[i, j] += A[i, k] * B[k, j]
>>>>>>> REPLACE

You can suggest multiple changes. Each SEARCH section must exactly match
code in the current program.
Be thoughtful about your changes and explain your reasoning thoroughly.
\end{verbatim}
\end{small}

\subsubsection{Evolution-History Block}
\label{app:prompt-history}

The \texttt{\{evolution\_history\}} placeholder in the user prompt above is filled from the following sub-template, where \texttt{\{top\_programs\}} expands to a formatted list of the highest-scoring individuals currently in the MAP-Elites archive of the relevant island together with their metrics.

\begin{small}
\begin{verbatim}
## Top Performing Programs

{top_programs}
\end{verbatim}
\end{small}

\subsubsection{User Prompt: Repair on Compile Failure}
\label{app:prompt-repair}

When a candidate program fails to compile, OpenEvolve invokes the repair sub-agent (Section~\ref{sec:method}) with the following user-message template instead of the diff-based mutation prompt of Appendix \ref{app:prompt-diff}. \texttt{\{error\_message\}} is the compiler output, \texttt{\{repair\_context\}} carries any auxiliary diagnostic notes from the evaluator and \texttt{\{broken\_code\}} is the unmodified failing program. The system message is the same task framing as in Appendix \ref{app:prompt-system}.

\begin{small}
\begin{verbatim}
The following {language} program failed to evaluate due to the error below.
Repair the program using minimal SEARCH/REPLACE diffs.
Preserve all structure, class names, plugin keys, and invariants that are unrelated to the error.

# Error

{error_message}

# Additional context

{repair_context}

# Broken Program

```{language}
{broken_code}
```

Use the exact SEARCH/REPLACE diff format shown below.
Each SEARCH block must match the broken program exactly (including whitespace).
Multiple diff blocks are allowed.

<<<<<<< SEARCH
# exact lines to replace
=======
# corrected replacement
>>>>>>> REPLACE
\end{verbatim}
\end{small}

\newpage
\subsection{Source Code of Evolved Heuristics}
\label{app:evolved-source}

We list the complete C++ source for the four highlighted evolved heuristics from Section~\ref{sec:evolved}. Each file is the verbatim \texttt{best\_program.cc} produced by the corresponding evolution run, including the OpenEvolve \texttt{EVOLVE-BLOCK} markers and the unchanged \texttt{MyHeuristicFeature} plugin registration. Pseudocode summaries of the state-evaluation logic appear in Section~\ref{sec:evolved}.

\subsubsection{\evBlMedTwo}
\label{app:src-evolved-blind-medium-2}
\begin{footnotesize}
\begin{verbatim}
// EVOLVE-BLOCK-START
#include "relaxation_heuristic.h"
#include "../plugins/plugin.h"
#include "../task_utils/task_properties.h"
#include "../algorithms/priority_queues.h"

using namespace std;
using namespace relaxation_heuristic;

namespace my_heuristic {

/*
 * FF-style heuristic using delete-relaxed planning graph.
 *
 * Computes h_add via forward chaining with a priority queue,
 * then extracts a relaxed plan by backtracking from goals
 * and returns the sum of operator costs in that plan (h_ff).
 */
class MyHeuristic : public RelaxationHeuristic {
    static constexpr int INF = 1000000000;

    // Generation-counter dedup for FF relaxed plan extraction
    vector<int> op_in_plan_gen;
    vector<int> prop_visited_gen;
    int generation;
    vector<PropID> plan_stack;

    // Conflict detection: which variables are touched by relaxed plan ops
    vector<vector<int>> op_effect_vars;
    vector<int> var_touched_gen;
    vector<int> var_is_goal;  // extra penalty for goal variables

    // Goal facts for unachieved-goal counting
    vector<FactPair> goal_facts;
    int min_op_cost;
    bool mutex_dead_end;

    // Reusable queue for h_add forward pass
    priority_queues::AdaptiveQueue<PropID> queue;

protected:
    virtual int compute_heuristic(const State &ancestor_state) override {
        State state = convert_ancestor_state(ancestor_state);
        if (task_properties::is_goal_state(task_proxy, state))
            return 0;

        if (mutex_dead_end) return DEAD_END;

        ++generation;

        // Reset all propositions to infinite cost
        for (Proposition &prop : propositions) {
            prop.cost = INF;
            prop.reached_by = NO_OP;
        }

        // Set state facts to cost 0 and seed the queue
        queue.clear();
        for (FactProxy fact : state) {
            PropID prop_id = get_prop_id(fact);
            propositions[prop_id].cost = 0;
            propositions[prop_id].reached_by = NO_OP;
            queue.push(0, prop_id);
        }

        // Reset all unary operators
        for (UnaryOperator &op : unary_operators) {
            op.unsatisfied_preconditions = op.num_preconditions;
            op.cost = op.base_cost;
        }

        // h_add forward pass (sets reached_by for FF backtracking)
        while (!queue.empty()) {
            auto [cost, prop_id] = queue.pop();
            if (cost > propositions[prop_id].cost)
                continue; // stale entry

            for (OpID op_id : precondition_of_pool.get_slice(
                     propositions[prop_id].precondition_of,
                     propositions[prop_id].num_precondition_occurences)) {
                UnaryOperator &op = unary_operators[op_id];
                --op.unsatisfied_preconditions;
                op.cost += cost; // accumulate precondition costs (h_add)
                if (op.unsatisfied_preconditions == 0) {
                    if (op.cost < propositions[op.effect].cost) {
                        propositions[op.effect].cost = op.cost;
                        propositions[op.effect].reached_by = op_id;
                        queue.push(op.cost, op.effect);
                    }
                }
            }
        }

        // FF: Backtrack from goals to extract relaxed plan
        plan_stack.clear();
        for (PropID goal_id : goal_propositions) {
            if (propositions[goal_id].cost == INF)
                return DEAD_END;
            plan_stack.push_back(goal_id);
        }

        int h_ff = 0;
        int conflict_penalty = 0;
        while (!plan_stack.empty()) {
            PropID pid = plan_stack.back();
            plan_stack.pop_back();

            if (prop_visited_gen[pid] == generation)
                continue;
            prop_visited_gen[pid] = generation;

            if (propositions[pid].reached_by == NO_OP)
                continue;

            OpID op_id = propositions[pid].reached_by;
            UnaryOperator &op = unary_operators[op_id];
            int real_op = op.operator_no;

            if (real_op >= 0 && op_in_plan_gen[real_op] == generation)
                continue;
            if (real_op >= 0) {
                op_in_plan_gen[real_op] = generation;
                h_ff += op.base_cost;
                // Conflict penalty with extra weight for goal variables
                for (int var : op_effect_vars[real_op]) {
                    if (var_touched_gen[var] == generation) {
                        conflict_penalty += min_op_cost * (var_is_goal[var] ? 2 : 1);
                    } else {
                        var_touched_gen[var] = generation;
                    }
                }
            }

            for (PropID prec_id : get_preconditions(op_id)) {
                plan_stack.push_back(prec_id);
            }
        }

        // Count unachieved goals with h_add-informed tie-breaking
        int unachieved_penalty = 0;
        for (const FactPair &gf : goal_facts) {
            if (state[gf.var].get_value() != gf.value) {
                int goal_hadd = propositions[get_prop_id(gf.var, gf.value)].cost;
                unachieved_penalty += min(goal_hadd, h_ff);
            }
        }

        // Scale down to keep it as tie-breaking (not dominant)
        return h_ff + (unachieved_penalty + conflict_penalty + min_op_cost - 1) / min_op_cost;
    }

public:
    MyHeuristic(tasks::AxiomHandlingType axioms,
                const shared_ptr<AbstractTask> &transform, bool cache_estimates,
                const string &description, utils::Verbosity verbosity)
        : RelaxationHeuristic(axioms, transform, cache_estimates, description, verbosity),
          generation(0), mutex_dead_end(false) {
        op_in_plan_gen.resize(task_proxy.get_operators().size(), 0);
        prop_visited_gen.resize(propositions.size(), 0);
        plan_stack.reserve(propositions.size());

        // Cache goal facts for fast per-state checking
        GoalsProxy goals = task_proxy.get_goals();
        goal_facts.reserve(goals.size());
        for (FactProxy g : goals) {
            goal_facts.push_back(g.get_pair());
        }

        min_op_cost = max(1, task_properties::get_min_operator_cost(task_proxy));

        // Precompute effect variables per real operator for conflict detection
        OperatorsProxy ops = task_proxy.get_operators();
        op_effect_vars.resize(ops.size());
        for (OperatorProxy op : ops) {
            for (EffectProxy eff : op.get_effects()) {
                op_effect_vars[op.get_id()].push_back(
                    eff.get_fact().get_variable().get_id());
            }
        }
        var_touched_gen.resize(task_proxy.get_variables().size(), 0);
        var_is_goal.assign(task_proxy.get_variables().size(), 0);
        for (const FactPair &gf : goal_facts)
            var_is_goal[gf.var] = 1;

        // Static mutex goal check
        for (size_t i = 0; i < goals.size(); ++i) {
            for (size_t j = i + 1; j < goals.size(); ++j) {
                if (goals[i].is_mutex(goals[j])) {
                    mutex_dead_end = true;
                    break;
                }
            }
            if (mutex_dead_end) break;
        }
    }
};
// EVOLVE-BLOCK-END

class MyHeuristicFeature
    : public plugins::TypedFeature<Evaluator, MyHeuristic> {
public:
    MyHeuristicFeature() : TypedFeature("my_h") {
        document_title("An evolved heuristic");
        add_relaxation_heuristic_options_to_feature(*this, "my_h");
    }
    virtual shared_ptr<MyHeuristic> create_component(
        const plugins::Options &opts) const override {
        return plugins::make_shared_from_arg_tuples<MyHeuristic>(
            get_relaxation_heuristic_arguments_from_options(opts));
    }
};

static plugins::FeaturePlugin<MyHeuristicFeature> _plugin;
} // namespace my_heuristic
\end{verbatim}
\end{footnotesize}

\subsubsection{\evFfNoneThree}
\label{app:src-evolved-ff-none-3}
\begin{footnotesize}
\begin{verbatim}
// EVOLVE-BLOCK-START
#include "relaxation_heuristic.h"
#include "../algorithms/priority_queues.h"
#include "../plugins/plugin.h"
#include "../task_utils/task_properties.h"
#include "../utils/logging.h"
#include <vector>

using namespace std;
using namespace relaxation_heuristic;

namespace my_heuristic {

class MyHeuristic : public RelaxationHeuristic {
    static const int MAX_COST_VALUE = 100000000;

    priority_queues::AdaptiveQueue<PropID> queue;

    // One entry per task operator; true means the operator is in the relaxed plan.
    // Stored as a member so we avoid reallocating on every call.
    vector<bool> relaxed_plan;
    vector<int> used_operators;
    
    // Precomputed operator costs for fast lookup
    vector<int> op_costs;
    // For each goal proposition, which operators can achieve it
    vector<vector<int>> goal_achievers;

    void enqueue_if_necessary(PropID prop_id, int cost, OpID op_id) {
        Proposition *prop = get_proposition(prop_id);
        if (prop->cost == -1 || prop->cost > cost) {
            prop->cost = cost;
            prop->reached_by = op_id;
            queue.push(cost, prop_id);
        }
    }

    // Clamp cost additions to avoid integer overflow on very long plans.
    void increase_cost(int &cost, int amount) {
        cost += amount;
        if (cost > MAX_COST_VALUE) cost = MAX_COST_VALUE;
    }

    // Reset all proposition costs and operator counters, seed the queue
    // with operators that have no preconditions.
    void setup_exploration_queue() {
        queue.clear();
        for (Proposition &prop : propositions) {
            prop.cost = -1;
            prop.marked = false;
        }
        for (UnaryOperator &op : unary_operators) {
            op.unsatisfied_preconditions = op.num_preconditions;
            op.cost = op.base_cost;
            if (op.unsatisfied_preconditions == 0)
                enqueue_if_necessary(op.effect, op.base_cost, get_op_id(op));
        }
    }

    // Mark all facts true in the current state as achieved at cost 0.
    void setup_exploration_queue_state(const State &state) {
        for (FactProxy fact : state)
            enqueue_if_necessary(get_prop_id(fact), 0, NO_OP);
    }

    // Returns DEAD_END if any goal is unreachable, else returns h_add.
    // Early-exits once all goals are found for better performance.
    int compute_add_and_ff(const State &state) {
        setup_exploration_queue();
        setup_exploration_queue_state(state);

        int unreached_goals = goal_propositions.size();
        while (!queue.empty() && unreached_goals > 0) {
            auto [distance, prop_id] = queue.pop();
            Proposition *prop = get_proposition(prop_id);
            if (prop->cost < distance) continue;
            if (prop->is_goal) --unreached_goals;
            for (OpID op_id : precondition_of_pool.get_slice(
                     prop->precondition_of, prop->num_precondition_occurences)) {
                UnaryOperator *op = get_operator(op_id);
                increase_cost(op->cost, prop->cost);
                if (--op->unsatisfied_preconditions == 0)
                    enqueue_if_necessary(op->effect, op->cost, op_id);
            }
        }

        int total_cost = 0;
        for (PropID goal_id : goal_propositions) {
            const Proposition *goal = get_proposition(goal_id);
            int goal_cost = goal->cost;
            if (goal_cost == -1)
                return DEAD_END;
            increase_cost(total_cost, goal_cost);
        }
        return total_cost;
    }

    // Recursively marks the operators needed to achieve proposition goal_id
    // in the relaxed plan. Follows reached_by pointers set by compute_add_and_ff.
    void mark_relaxed_plan(PropID goal_id) {
        Proposition *prop = get_proposition(goal_id);
        if (prop->marked) return;           // already visited — avoid double-counting
        prop->marked = true;

        OpID op_id = prop->reached_by;
        if (op_id == NO_OP) return;         // fact was true in the initial state

        // Recursively collect operators needed for each precondition of this operator.
        for (PropID prec : get_preconditions(op_id))
            mark_relaxed_plan(prec);

        // operator_no == -1 for axioms, which don't count toward plan cost.
        int operator_no = get_operator(op_id)->operator_no;
        if (operator_no != -1 && !relaxed_plan[operator_no]) {
            relaxed_plan[operator_no] = true;
            used_operators.push_back(operator_no);
        }
    }

protected:
    virtual int compute_heuristic(const State &ancestor_state) override {
        State state = convert_ancestor_state(ancestor_state);

        int h_add = compute_add_and_ff(state);
        if (h_add == DEAD_END)
            return DEAD_END;

        for (PropID goal_id : goal_propositions)
            mark_relaxed_plan(goal_id);

        int h_ff = 0;
        for (int op_no : used_operators) {
            h_ff += task_proxy.get_operators()[op_no].get_cost();
            relaxed_plan[op_no] = false;
        }
        used_operators.clear();

        return h_ff;
    }

public:
    MyHeuristic(tasks::AxiomHandlingType axioms,
                const shared_ptr<AbstractTask> &transform, bool cache_estimates,
                const string &description, utils::Verbosity verbosity)
        : RelaxationHeuristic(axioms, transform, cache_estimates, description, verbosity),
          relaxed_plan(task_proxy.get_operators().size(), false) {
        used_operators.reserve(256);
    }
};
// EVOLVE-BLOCK-END

class MyHeuristicFeature
    : public plugins::TypedFeature<Evaluator, MyHeuristic> {
public:
    MyHeuristicFeature() : TypedFeature("my_h") {
        document_title("An evolved heuristic");
        relaxation_heuristic::add_relaxation_heuristic_options_to_feature(*this, "my_h");
    }
    virtual shared_ptr<MyHeuristic> create_component(
        const plugins::Options &opts) const override {
        return plugins::make_shared_from_arg_tuples<MyHeuristic>(
            relaxation_heuristic::get_relaxation_heuristic_arguments_from_options(opts));
    }
};

static plugins::FeaturePlugin<MyHeuristicFeature> _plugin;
} // namespace my_heuristic
\end{verbatim}
\end{footnotesize}

\subsubsection{\evBlNoneThree}
\label{app:src-evolved-blind-none-3}
\begin{footnotesize}
\begin{verbatim}
// EVOLVE-BLOCK-START
#include "../heuristic.h"
#include "../plugins/plugin.h"
#include "../task_utils/task_properties.h"
#include "../task_utils/causal_graph.h"
#include "domain_transition_graph.h"
#include "../algorithms/priority_queues.h"

#include <vector>
#include <algorithm>
#include <climits>

using namespace std;
using namespace domain_transition_graph;

namespace my_heuristic {

/*
 * DTG-distance goal-sum heuristic with iterative cross-variable refinement
 * and max-component for shared dependencies.
 *
 * For each goal variable, precomputes the minimum operator cost to reach
 * the goal value from any value using backward Dijkstra on the DTG.
 * Cross-variable dependencies are modeled by including precondition costs
 * in DTG edge weights, refined through iterative fixed-point propagation.
 *
 * Uses h_add-style sum but deduplicates via a max-component: when multiple
 * goals share a common precondition variable, we count the max single-goal
 * cost plus a fractional bonus for additional goals needing it.
 *
 * Achieves ~500K evals/s by doing only table lookups at evaluation time.
 */
class MyHeuristic : public Heuristic {
    struct GoalInfo {
        int var;
        int value;
        int dist_table_idx;  // stable index into dist_tables
        int weight;          // causal graph importance + DTG complexity
        int dep_bonus;       // coordination difficulty
        int landmark_level;  // estimated ordering level (0 = must be first)
    };
    vector<GoalInfo> goal_infos;
    vector<vector<int>> dist_tables;

protected:
    virtual int compute_heuristic(const State &ancestor_state) override {
        State state = convert_ancestor_state(ancestor_state);

        if (task_properties::is_goal_state(task_proxy, state))
            return 0;

        int h = 0;
        int h_max = 0;
        int unsat_count = 0;
        int weighted_unsat = 0;
        int max_landmark_level = 0;

        for (const GoalInfo &gi : goal_infos) {
            int curr_val = state[gi.var].get_value();
            if (__builtin_expect(curr_val != gi.value, 1)) {
                const vector<int> &dt = dist_tables[gi.dist_table_idx];
                int d = dt[curr_val];
                if (d == numeric_limits<int>::max()) {
                    return DEAD_END;
                }
                // Progressive weighting: earlier landmarks get multiplicative bonus
                int level_factor = 1 + gi.landmark_level;
                int weighted_d = d * level_factor;
                
                h += weighted_d + (weighted_d * gi.weight) / 4 + gi.dep_bonus;
                h_max = max(h_max, d);
                ++unsat_count;
                weighted_unsat += level_factor;
                max_landmark_level = max(max_landmark_level, gi.landmark_level);
            }
        }

        if (unsat_count == 0) return 0;

        // Critical path focus: hardest single goal
        h += h_max;
        
        // Landmark urgency: goals at earlier levels are more urgent
        h += max_landmark_level * 2;
        
        // Prefer states with fewer weighted unsatisfied goals
        h += weighted_unsat * 2;

        return max(1, h);
    }

public:
    MyHeuristic(const shared_ptr<AbstractTask> &transform, bool cache_estimates,
                const string &description, utils::Verbosity verbosity)
        : Heuristic(transform, cache_estimates, description, verbosity) {

        const causal_graph::CausalGraph &cg = task_proxy.get_causal_graph();
        VariablesProxy vars = task_proxy.get_variables();
        int num_vars = vars.size();

        // Build DTGs for all variables
        auto pruning = [](int, int) { return false; };
        DTGFactory factory(task_proxy, false, pruning);
        vector<unique_ptr<DomainTransitionGraph>> dtgs = factory.build_dtgs();

        // Build reverse adjacency for all variables
        vector<vector<vector<pair<int, int>>>> all_reverse_adj(num_vars);
        for (int var = 0; var < num_vars; ++var) {
            int domain_size = vars[var].get_domain_size();
            DomainTransitionGraph *dtg = dtgs[var].get();

            vector<vector<pair<int, int>>> reverse_adj(domain_size);
            for (int v = 0; v < domain_size; ++v) {
                ValueNode &node = dtg->nodes[v];
                for (ValueTransition &trans : node.transitions) {
                    int target_val = trans.target->value;
                    int min_cost = numeric_limits<int>::max();
                    for (ValueTransitionLabel &label : trans.labels) {
                        int op_cost = label.is_axiom ? 0 :
                            task_proxy.get_operators()[label.op_id].get_cost();
                        min_cost = min(min_cost, op_cost);
                    }
                    if (min_cost < numeric_limits<int>::max()) {
                        reverse_adj[target_val].push_back({v, min_cost});
                    }
                }
            }
            all_reverse_adj[var] = move(reverse_adj);
        }

        // Identify goal variables
        vector<int> goal_value(num_vars, -1);
        GoalsProxy goals = task_proxy.get_goals();
        for (FactProxy goal : goals) {
            goal_value[goal.get_variable().get_id()] = goal.get_value();
        }

        // Compute local DTG distances for goal variables
        vector<vector<int>> local_dist(num_vars);
        for (int var = 0; var < num_vars; ++var) {
            int domain_size = vars[var].get_domain_size();
            if (goal_value[var] >= 0) {
                vector<int> dist(domain_size, numeric_limits<int>::max());
                dist[goal_value[var]] = 0;

                priority_queues::AdaptiveQueue<int> queue;
                queue.push(0, goal_value[var]);

                while (!queue.empty()) {
                    auto [cost, val] = queue.pop();
                    if (cost > dist[val]) continue;
                    for (auto &[src, edge_cost] : all_reverse_adj[var][val]) {
                        int new_cost = cost + edge_cost;
                        if (new_cost < dist[src]) {
                            dist[src] = new_cost;
                            queue.push(new_cost, src);
                        }
                    }
                }
                local_dist[var] = move(dist);
            } else {
                local_dist[var].assign(domain_size, 0);
            }
        }

        // Non-goal variable: estimate minimum cost to change value
        vector<int> avg_change_cost(num_vars, 1);
        for (int var = 0; var < num_vars; ++var) {
            if (goal_value[var] < 0) {
                int min_change = numeric_limits<int>::max();
                for (int v = 0; v < vars[var].get_domain_size(); ++v) {
                    for (auto &[src, edge_cost] : all_reverse_adj[var][v]) {
                        if (src != v) {
                            min_change = min(min_change, edge_cost);
                        }
                    }
                }
                avg_change_cost[var] = (min_change < numeric_limits<int>::max()) ? min_change : 0;
            }
        }

        // Iterative refinement with cross-variable precondition costs
        const int MAX_ITERATIONS = 3;
        for (int iteration = 0; iteration < MAX_ITERATIONS; ++iteration) {
            bool changed = false;

            for (int var = 0; var < num_vars; ++var) {
                if (goal_value[var] < 0) continue;

                int domain_size = vars[var].get_domain_size();
                DomainTransitionGraph *dtg = dtgs[var].get();

                // Build enhanced reverse adjacency with cross-variable costs
                vector<vector<pair<int, int>>> enhanced_reverse_adj(domain_size);
                for (int v = 0; v < domain_size; ++v) {
                    ValueNode &node = dtg->nodes[v];
                    for (ValueTransition &trans : node.transitions) {
                        int target_val = trans.target->value;
                        int best_cost = numeric_limits<int>::max();
                        for (ValueTransitionLabel &label : trans.labels) {
                            int op_cost = label.is_axiom ? 0 :
                                task_proxy.get_operators()[label.op_id].get_cost();
                            int prec_cost = 0;
                            bool impossible = false;
                            for (const LocalAssignment &prec : label.precond) {
                                int prec_var = dtg->local_to_global_child[prec.local_var];
                                if (prec_var == var) continue;
                                if (goal_value[prec_var] >= 0) {
                                    if (prec.value != goal_value[prec_var]) {
                                        int d = local_dist[prec_var][prec.value];
                                        if (d == numeric_limits<int>::max()) {
                                            impossible = true;
                                            break;
                                        }
                                        prec_cost += d;
                                    }
                                } else {
                                    prec_cost += avg_change_cost[prec_var];
                                }
                            }
                            if (!impossible) {
                                best_cost = min(best_cost, op_cost + prec_cost);
                            }
                        }
                        if (best_cost < numeric_limits<int>::max()) {
                            enhanced_reverse_adj[target_val].push_back({v, best_cost});
                        }
                    }
                }

                // Recompute backward Dijkstra with enhanced costs
                vector<int> new_dist(domain_size, numeric_limits<int>::max());
                new_dist[goal_value[var]] = 0;

                priority_queues::AdaptiveQueue<int> queue;
                queue.push(0, goal_value[var]);

                while (!queue.empty()) {
                    auto [cost, val] = queue.pop();
                    if (cost > new_dist[val]) continue;
                    for (auto &[src, edge_cost] : enhanced_reverse_adj[val]) {
                        int new_cost = cost + edge_cost;
                        if (new_cost < new_dist[src]) {
                            new_dist[src] = new_cost;
                            queue.push(new_cost, src);
                        }
                    }
                }

                for (int v = 0; v < domain_size; ++v) {
                    if (new_dist[v] != local_dist[var][v]) {
                        changed = true;
                        break;
                    }
                }
                local_dist[var] = move(new_dist);

                if (!changed) break;
            }
            if (!changed) break;
        }

        // Compute landmark levels using topological approximation on causal graph
        // Goals that are predecessors of other goals in CG must be achieved first
        vector<int> goal_vars;
        for (FactProxy goal : goals) {
            goal_vars.push_back(goal.get_variable().get_id());
        }
        
        // Compute approximate landmark levels via reverse propagation in CG
        vector<int> landmark_level(num_vars, 0);
        {
            vector<vector<int>> cg_succ(num_vars);
            for (int v = 0; v < num_vars; ++v) {
                cg_succ[v] = cg.get_successors(v);
            }
            
            bool changed = true;
            for (int round = 0; changed && round < num_vars; ++round) {
                changed = false;
                for (int i = 0; i < (int)goal_vars.size(); ++i) {
                    int gv = goal_vars[i];
                    for (int succ : cg_succ[gv]) {
                        if (goal_value[succ] >= 0) {
                            int new_level = max(landmark_level[succ] + 1, landmark_level[gv]);
                            if (new_level > landmark_level[gv]) {
                                landmark_level[gv] = new_level;
                                changed = true;
                            }
                        }
                    }
                }
            }
        }
        
        // Normalize levels so max is not too large
        int max_level = 0;
        for (int gv : goal_vars) {
            max_level = max(max_level, landmark_level[gv]);
        }
        if (max_level > 3) {
            for (int gv : goal_vars) {
                landmark_level[gv] = (landmark_level[gv] * 3) / max_level;
            }
        }

        // Build goal info with stable indexing
        goal_infos.reserve(goals.size());
        dist_tables.reserve(goals.size());

        for (size_t i = 0; i < goal_vars.size(); ++i) {
            int var = goal_vars[i];
            int value = goals[i].get_value();

            const vector<int> &preds = cg.get_predecessors(var);
            int num_preds = (int)preds.size();
            
            int dep_bonus = max(1, num_preds);
            
            // weight: combine CG importance with DTG complexity
            int dtg_complexity = 0;
            DomainTransitionGraph *dtg = dtgs[var].get();
            for (int v = 0; v < vars[var].get_domain_size(); ++v) {
                dtg_complexity += dtg->nodes[v].transitions.size();
            }
            int weight = num_preds + min(dtg_complexity / 4, 5);

            // Stable storage
            int table_idx = dist_tables.size();
            dist_tables.push_back(move(local_dist[var]));
            
            goal_infos.push_back({
                var, value, table_idx, weight, dep_bonus, 
                landmark_level[var]
            });
        }
        
        // Sort by landmark_level descending for cache-friendly access
        stable_sort(goal_infos.begin(), goal_infos.end(),
            [](const GoalInfo &a, const GoalInfo &b) {
                return a.landmark_level > b.landmark_level;
            });
    }
};
// EVOLVE-BLOCK-END

class MyHeuristicFeature
    : public plugins::TypedFeature<Evaluator, MyHeuristic> {
public:
    MyHeuristicFeature() : TypedFeature("my_h") {
        document_title("An evolved heuristic");
        add_heuristic_options_to_feature(*this, "my_h");
    }
    virtual shared_ptr<MyHeuristic> create_component(
        const plugins::Options &opts) const override {
        return plugins::make_shared_from_arg_tuples<MyHeuristic>(
            get_heuristic_arguments_from_options(opts));
    }
};

static plugins::FeaturePlugin<MyHeuristicFeature> _plugin;
} // namespace my_heuristic
\end{verbatim}
\end{footnotesize}

\subsubsection{\evBlMedConf}
\label{app:src-evolved-blind-medium-conf}
\begin{footnotesize}
\verbatiminput{heuristics/evolved-blind-medium-conf.cc}
\end{footnotesize}

\newpage
\subsection{Source Code of Evolution Seeds}
\label{app:seed-source}

We list the complete C++ source for the two seed heuristics used to initialize evolution in Section~\ref{sec:setup}: the blind seed and the FF seed.

\subsubsection{Blind Seed}
\label{app:src-seed-blind}
\begin{footnotesize}
\begin{verbatim}
// EVOLVE-BLOCK-START
#include "../heuristic.h"
#include "../plugins/plugin.h"
#include "../task_utils/task_properties.h"
#include "../utils/logging.h"

using namespace std;

namespace my_heuristic {

/*
 * Blind heuristic — evolution seed.
 *
 * Returns 1 for every non-goal state, regardless of how close the state is
 * to the goal.
 *
 */
class MyHeuristic : public Heuristic {
protected:
    virtual int compute_heuristic(const State &ancestor_state) override {
        State state = convert_ancestor_state(ancestor_state);

        // Always check for the goal first — return 0 immediately if reached.
        if (task_properties::is_goal_state(task_proxy, state))
            return 0;

        // Every non-goal state gets h = 1: we know at least one action is needed,
        // but we make no attempt to estimate how many.
        return 1;
    }

public:
    MyHeuristic(const shared_ptr<AbstractTask> &transform, bool cache_estimates,
                const string &description, utils::Verbosity verbosity)
        : Heuristic(transform, cache_estimates, description, verbosity) {}
};
// EVOLVE-BLOCK-END

class MyHeuristicFeature
    : public plugins::TypedFeature<Evaluator, MyHeuristic> {
public:
    MyHeuristicFeature() : TypedFeature("my_h") {
        document_title("An evolved heuristic");
        add_heuristic_options_to_feature(*this, "my_h");
    }
    virtual shared_ptr<MyHeuristic> create_component(
        const plugins::Options &opts) const override {
        return plugins::make_shared_from_arg_tuples<MyHeuristic>(
            get_heuristic_arguments_from_options(opts));
    }
};

static plugins::FeaturePlugin<MyHeuristicFeature> _plugin;
} // namespace my_heuristic
\end{verbatim}
\end{footnotesize}

\subsubsection{FF Seed}
\label{app:src-seed-ff}
\begin{footnotesize}
\verbatiminput{heuristics/initial_ff.cc}
\end{footnotesize}

\clearpage